\documentclass[sigconf]{acmart}
\pdfoutput=1 
\renewcommand\footnotetextcopyrightpermission[1]{} 
\usepackage{booktabs} 

\usepackage{algorithm}
\usepackage[noend]{algorithmic}
\usepackage{siunitx}

\newcommand{\B}{{\mathbb{B}}}

\newcommand{\N}{{\mathbb{N}}}

\newcommand{\R}{{\mathbb{R}}}

\newcommand{\W}{{\mathcal{W}}}

\newcommand{\argmin}{\textrm{arg}\min}

\DeclareMathOperator{\atantwo}{atan2}
\renewcommand{\matrix}[1]{\begin{bmatrix}#1\end{bmatrix}}

\usepackage{amssymb,latexsym,amsfonts,amsmath}
\newtheorem{theorem}{Theorem}[section] 
\newtheorem{lemma}[theorem]{Lemma}

\newtheorem{problem}[theorem]{Problem}

\usepackage{color}

\setcopyright{none}

\begin{document}
\title{Formal Verification of Neural Network Controlled \\Autonomous Systems}


%
\author{Xiaowu Sun $\qquad\qquad\qquad$ Haitham Khedr $\qquad\qquad\qquad$ Yasser Shoukry}
\affiliation{%
  \institution{Department of Electrical and Computer Engineering \\ University of Maryland, College Park}
}
\email{{xsun24,hkhedr}@umd.edu, yshoukry@ece.umd.edu}

%
%
%
%
%

\renewcommand{\shortauthors}{X. Sun et al.}

\begin{abstract}
In this paper, we consider the problem of formally verifying the safety of an autonomous robot equipped with a Neural Network (NN) controller that processes LiDAR images to produce control actions. Given a workspace that is characterized by a set of polytopic obstacles, our objective is to compute the set of safe initial conditions such that a robot trajectory starting from these initial conditions is guaranteed to avoid the obstacles. Our approach is to construct a finite state abstraction of the system and use standard reachability analysis over the finite state abstraction to compute the set of the safe initial states. The first technical problem in computing the finite state abstraction is to mathematically model the imaging function that maps the robot position to the LiDAR image. To that end, we introduce the notion of imaging-adapted sets as partitions of the workspace in which the imaging function is guaranteed to be affine. Based on this notion, and resting on efficient algorithms in the literature of computational geometry, we develop a polynomial-time algorithm to partition the workspace into imaging-adapted sets along with computing the corresponding affine imaging functions. Given this workspace partitioning, a discrete-time linear dynamics of the robot, and a pre-trained NN controller with Rectified Linear Unit (ReLU) nonlinearity, the second technical challenge is to analyze the behavior of the neural network. To that end, and thanks to the ReLU functions being piecewise linear functions, we utilize a Satisfiability Modulo Convex (SMC) encoding to enumerate all the possible segments of different ReLUs. SMC solvers then use a Boolean satisfiability solver and a convex programming solver and decompose the problem into smaller subproblems. At each iteration, the  Boolean satisfiability solver searches for a candidate assignment for the different ReLU segments while completely abstracting the robot dynamics. Convex programming is then used to check the feasibility of the proposed ReLU phases against the dynamic and imagining constraints, or generate succinct explanations for their infeasibility to reduce the search space. To accelerate this process, we develop a pre-processing algorithm that could rapidly prune the space feasible ReLU segments. Finally, we demonstrate the efficiency of the proposed algorithms using numerical simulations with increasing complexity of the neural network controller.
\end{abstract}

%
%
%


\maketitle

\section{Introduction}
From simple logical constructs to complex deep neural network models, Artificial Intelligence (AI)-agents are increasingly controlling physical/mechanical systems. Self-driving cars, drones, and smart cities are just examples of such systems to name a few. However, regardless of the explosion in the use of AI within a multitude of cyber-physical systems (CPS) domains, the safety and reliability of these AI-enabled CPS is still an under-studied problem. It is then unsurprising the failure of these AI-controlled CPS in several, safety-critical, situations leading to human fatalities~\cite{AccidentWiki}. 


Motivated by the urgency to study the safety, reliability, and potential problems that can rise and impact the society by the deployment of AI-enabled systems in the real world, several works in the literature focused on the problem of designing deep neural networks that are robust to the so-called adversarial examples~\cite{ferdowsi2018robust,everitt2018agi,charikar2017learning,steinhardt2017certified,munoz2017towards,paudice2018label,ruan2018global}. Unfortunately, these techniques focus mainly on the robustness of the learning algorithm with respect to data outliers without providing guarantees in terms of safety and reliability of the decisions taken by these neural networks. To circumvent this drawback, and motivated by the wealth of adversarial example generation approaches for neural networks, recent works focused on three main techniques namely (i) testing of neural networks, (ii) falsification (semi-formal verification) of neural networks, and (iii) formal verification of neural networks.

Representatives of the first class, namely testing of neural networks, are the work reported in~\cite{pei2017deepxplore,tian2017deeptest,wicker2018feature,YouchengTesting2018,LeiDeepGauge2018,Wang2018Testing,LeiDeepMutation2018,srisakaokul2018multiple,MengshiDeepRoad2018,YouchengConcolic2018} in which the neural network is treated as a white box, and test cases are generated to maximize different coverage criteria. Such coverage criteria include neuron coverage, condition/decision coverage, and multi-granularity testing criteria. On the one hand, maximizing test coverage give system designers confidence that the networks are reasonably free from defect. On the other hand, testing do not formally guarantee that a neural network satisfy a formal specification.  

Unfortunately, testing techniques focuses mainly on the neural network as a component without taking into consideration the effect of its decisions on the entire system behavior.
This motivated researchers to focus on falsification (or semi-formal verification) of autonomous systems that include machine learning components~\cite{dreossi2017compositional,tuncali2018simulation,zhang2018two}. In such falsification frameworks, the objective is to generate corner test cases that will lead the whole system to violate a system-level specification. To that end, advanced 3D models and image environments are used to bridge the gap between the virtual world and the real world. By parametrizing the input to these 3D models (e.g., position of objects, position of light sources, intensity of light sources) and sampling the parameter space in a fashion that maximizes the falsification of the safety property, falsification frameworks can simulate several test cases until a counterexample is found~\cite{dreossi2017compositional,tuncali2018simulation,zhang2018two}.

While testing and falsification frameworks are powerful tools to find corner cases in which the neural network or the neural network enabled system will fail, they lack the rigor promised by formal verification methods. Therefore, several researchers pointed to the urgent need of using formal methods to verify the behavior of neural networks and neural network enabled system~\cite{kurd2003establishing,seshia2016towards,seshia2018formal,leikeAIsafety2017,leofante2018automated,scheibler2015towards}. As a result, recent works in the literature attempted the problem of applying formal verification techniques to neural network models. 

Applying formal verification to neural network models comes with its unique challenges. First and foremost is the lack of widely-accepted, precise, mathematical specifications capturing the correct behavior of a neural network. Therefore, recent works focused entirely on verifying neural networks against simple input-output specifications~\cite{katz2017reluplex,ehlers2017formal,bunel2018unified,ruan2018reachability,dutta2018output,pulina2010abstraction}. Such input-output techniques compute a guaranteed range for the output of a deep neural network given a set of inputs represented as a convex polyhedron.  
To that end, several algorithms that takes advantage of the piecewise linear nature of the Rectified Linear Unit (ReLU) activation functions (one of the most famous nonlinear activation functions in deep neural networks) have been proposed. For example, by using binary variables to encode piecewise linear functions, the constraints of ReLU functions are encoded as a Mixed-Integer Linear Programming (MILP). Combining output specifications that are expressed in terms of Linear Programming (LP), the verification problem eventually turns to a MILP feasibility problem~\cite{dutta2018output,tjeng2017verifying}. 

Using off-the-shelf MILP solvers does not lead to a scalable approach to handle neural networks with hundreds and thousands of neurons~\cite{ehlers2017formal}. To circumvent this problem, several MILP-like solvers targeted toward the neural network verification problem are proposed. For example, the work reported in~\cite{katz2017reluplex} proposed a modified Simplex algorithm (originally used to solve linear programs) to take into account ReLU nonlinearities as well. Similarly, the work reported in~\cite{ehlers2017formal} combines a Boolean satisfiability solving along with a linear over-approximation of piecewise linear functions to verify ReLU neural networks against convex specifications. Other techniques that exploit specific geometric structures of the specifications are also proposed~\cite{gehr2018ai,xiang2017reachable}. A thorough survey on different algorithms for verification of neural networks against input-output range specifications can be found in~\cite{xiang2017survey} and the references within.

Unfortunately, the input-output range properties, studied so far in the literature, are simplistic and fails to capture the safety and reliability of cyber-physical systems when controlled by a neural network. Therefore, in this paper, we focus instead on the problem of formal verification of a neural network controlled robot against system-level safety specifications. In particular, we consider the problem in which a robot utilizes a LiDAR scanner to sense its environment. The LiDAR image is then processed by a neural network controller to compute the control inputs. Such scenario is common in the literature of behavioral cloning and imitation control in which the neural network is trained to imitate the actions generated by experts controlling the robot~\cite{kahn2017plato}. With the objective to verify the safety of these robots, we develop a framework that can take into account the robot continuous dynamics, the workspace configuration, the LiDAR imaging, and the neural network, and compute the set of initial states for the robot that is guaranteed to produce robot trajectories that are safe and collision free.

To carry out the prescribed formal verification problem, we need a mathematical model that captures the LiDAR imaging process. This is the process that generates the LiDAR images based on the robot pose relative to the workspace objects.  Therefore, the first contribution of this paper is to introduce the notion of imaging-adapted sets. These are workspace partitions within which the map between robot pose and LiDAR images are mathematically captured by an affine map. Given this notion, and thanks to the advances in the literature of computational graphics, we develop a polynomial-time algorithm that can partition the workspace into imaging-adapted sets along with the corresponding affine maps.



Given the partitioned workspace along with a pre-trained neural network and the robot dynamics, we compute a finite state abstraction of the closed loop system that enjoys a simulation relation with the original system. The main challenge in computing this finite state abstraction is to analyze the behavior of the neural network controller. Similar to previous works in the literature, we strict our focus to neural networks with Rectified Linear Unit (ReLU) nonlinearities and we develop a Satisfiability Modulo Convex (SMC) programming algorithm that uses a combination of a Boolean satisfiability solver and a convex programming solver to iteratively reason about the neural network nonlinearity along with the dynamics and the imaging constraints. At each iteration, the boolean satisfiability solver searches for a candidate assignment determining whether ReLU units are active while ignoring the neural network weights, the robot dynamics, and the LiDAR imaging. The convex programming solver is then used to check the feasibility of the proposed ReLU assignment against the neural network weights, the robot dynamics, and the LiDAR imaging. If the ReLU assignment is deemed infeasible, then the SMC solver will generate succinct explanations for their infeasibility to reduce the search space. To accelerate the process, we develop a pre-processing algorithm that can reduce the space of ReLU assignments.

Once the finite state abstraction is computed, we use standard reachability analysis techniques to compute the set of safe initial states.
To summarize, the contributions of this paper can be summarized as follows:\\
\textbf{1-} A framework for formally proving safety properties of autonomous robots controlled by neural network controllers that process LiDAR images to compute control inputs.\\
\textbf{2-} A notion of imaging-adapted sets along with a polynomial-time algorithm for partitioning the workspace into such sets while computing an affine model capturing the LiDAR imaging process. \\
\textbf{3-} An SMC-based algorithm combined with an SMC-based pre-processing for computing finite abstractions of the neural network controlled autonomous robot.


\section{Problem Formulation}

\begin{figure}[!ht]
    \includegraphics[width=0.45\textwidth]{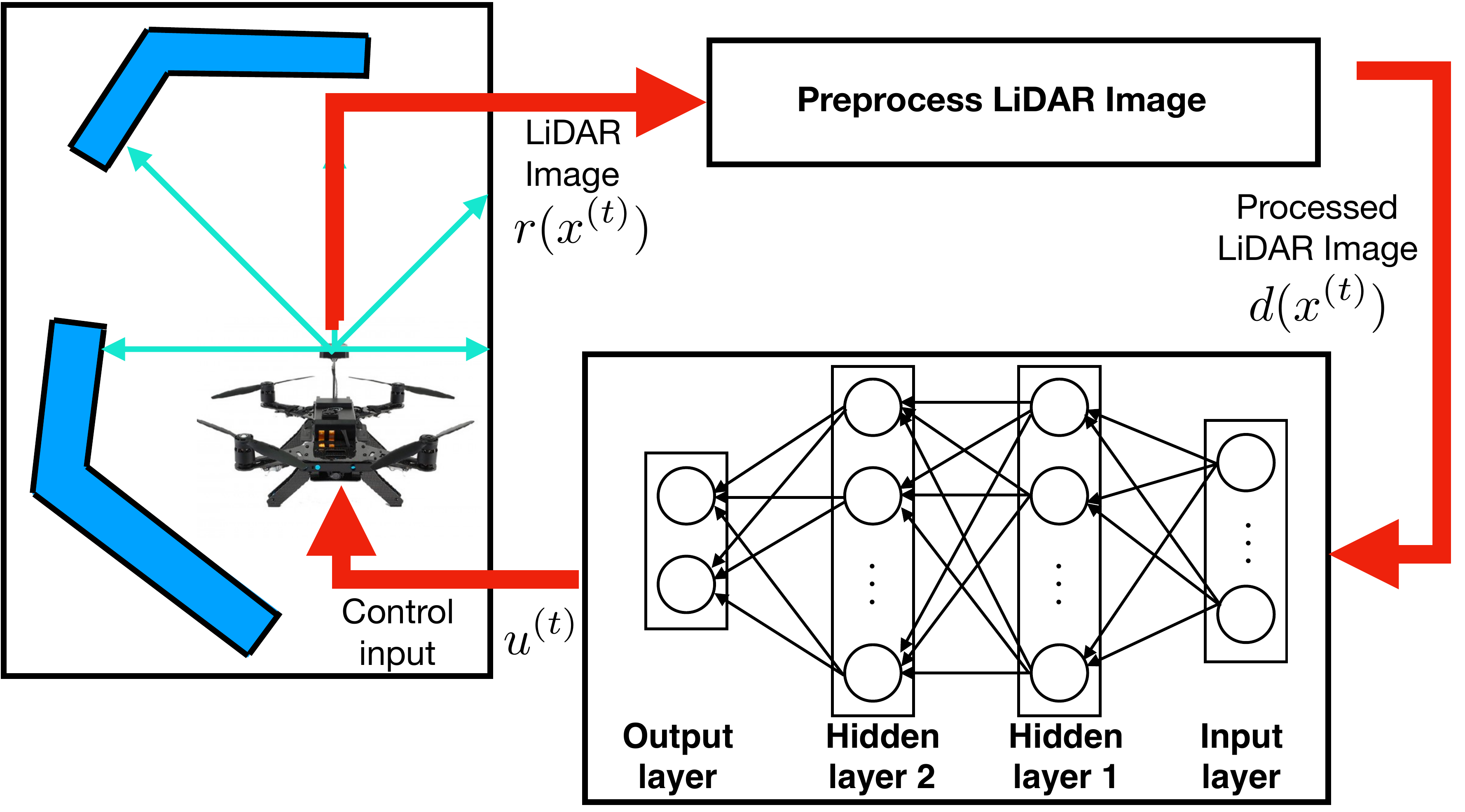}\vspace{-3mm}
    \caption{Pictorial representation of the problem setup under consideration.}
    \label{fig:formulation}
    \vspace{-6mm}
\end{figure}

\subsection{Notation}
The symbols $\N$, $\R, \R^{+}$ and $\B$ denote the set of natural, real, positive real, and Boolean numbers, respectively. The symbols $\land, \lnot$ and $\to$ denote the logical AND, logical NOT, and logical IMPLIES operators, respectively. Given two real-valued vectors $x_1 \in \R^{n_1}$ and $x_2 \in \R^{n_2}$, we denote by $(x_1, x_2) \in \R^{n_1 + n_2}$ the column vector $[x_1^T, x_2^T]$. Similarly, for a vector $x \in \R^n$, we denote by $x_i \in \R$ the $i$th element of $x$. For two vectors $x_1, x_2 \in \R^n$, we denote by $\max(x_1,x_2)$ the element-wise maximum. 
For a set $S \subset \R^{n}$, we denote the boundary and the interior of this set by $\partial S$ and $\text{int}(S)$, respectively.
Given two sets $S_1$ and $S_2$, $f: S_1 \rightrightarrows S_2$ and $f: S_1 \rightarrow S_2$ denote a set-valued and ordinary map, respectively. Finally, given a vector $z = (x,y) \in \R^2$, we denote by $\atantwo(z) = \atantwo(y,x)$.

\subsection{Dynamics and Workspace}
We consider an autonomous robot moving in a 2-dimensional polytopic (compact and convex) workspace $\W \subset \R^2$. 
We assume that the robot must avoid the workspace boundaries $\partial W$ along with a set of obstacles $\{\mathcal{O}_1, \ldots, \mathcal{O}_o\}$,  with $\mathcal{O}_i \subset \W$ which is assumed to be polytopic. We denote by $\mathcal{O}$ the set of the obstacles and the workspace boundaries which needs to be avoided, i.e., $\mathcal{O} = \{\partial W, \mathcal{O}_1, \ldots, \mathcal{O}_o\}$.
The dynamics of the robot is described by a discrete-time linear system of the form:
\begin{equation}
    \label{eq:dyn}    
    x^{(t+1)} = A x^{(t)} + B u^{(t)},
\end{equation}

\noindent where $x^{(t)} \in \mathcal{X} \subseteq \R^{n}$ is the state of robot at time $t \in \N$ and 
$u^{(t)} \subseteq \R^{m}$ is the robot input.
The matrices $A$ and $B$ represent the robot dynamics and have appropriate dimensions. For a robot with nonlinear dynamics that is either differentially flat or feedback linearizable,  the state space model~\eqref{eq:dyn} corresponds to its feedback linearized dynamics. We denote by $\zeta(x) \in \R^2$ the natural projection of $x$ onto the workspace $\W$, i.e., $\zeta(x^{(t)})$ is the position of the robot at time $t$.


%


\subsection{LiDAR Imaging}
We consider the case when the autonomous robot uses a LiDAR scanner to sense its environment. 
The LiDAR scanner emits a set of $N$ lasers evenly distributed in a $2\pi$ degree fan. We denote by $\theta_{\text{lidar}}^{(t)} \in \R$ the heading angle of the LiDAR at time $t$. Similarly, we denote by $\theta_{i}^{(t)} = \theta_{\text{lidar}}^{(t)} + (i - 1) \frac{2\pi}{N}$, with $i \in \{1,\ldots, N\}$, the angle of the $i$th laser beam at time $t$ where $\theta_{1}^{(t)} = \theta_{\text{lidar}}^{(t)}$ and by $\theta^{(t)} = (\theta_1^{(t)}, \ldots, \theta_N^{(t)})$ the vector of the angles of all the laser beams. While the heading angle of the LiDAR, $\theta_{\text{lidar}}^{(t)}$, changes as the robot poses changes over time, i.e.:
$$ \theta_{\text{lidar}}^{(t)} = f(x^{(t)}) $$
for some nonlinear function $f$, in this paper we focus on the case when the heading angle of the LiDAR, $\theta_{\text{lidar}}^{(t)}$, is fixed over time and we will drop the superscript $t$ from the notation. Such condition is satisfied in many real-world scenarios whenever the robot is moving while maintaining a fixed pose (e.g. a Quadrotor whose yaw angle is maintained constant).

For the $i$th laser beam, the observation signal $r_i(x^{(t)}) \in \R$ is the distance measured between the robot position $\zeta(x^{(t)})$ and the nearest obstacle in the $\theta_i$ direction, i.e.:
\begin{align}
 r_i(x^{(t)}) &= \min_{\mathcal{O}_i \in \mathcal{O}} \min_{z \in \mathcal{O}_i} \Vert z - \zeta(x^{(t)}) \Vert_2 
\nonumber\\  & \qquad\text{s.t.} \quad \atantwo(z - \zeta(x^{(t)})) = \theta_i.
 \label{eq:image_func_nonlinear}
\end{align}
The final LiDAR image $d(x^{(t)}) \in \R^{2N}$ is generated by processing the observations $r(x^{(t)})$ as follows:
\begin{align}
    d_i(x^{(t)}) &= \left(r_i(x^{(t)}) \cos \theta_i, \; r_i(x^{(t)}) \sin \theta_i \right), \nonumber\\ 
    d(x^{(t)}) &= \left(d_1(x^{(t)}), \ldots d_N(x^{(t)}) \right).
        \label{eq:image}
\end{align}



\subsection{Neural Network Controller}
We consider a pre-trained neural network controller $f_{\text{NN}}: \R^{2N}\to \R^m$ that processes the LiDAR images to produce control actions with $L$ internal and fully connected layers in addition to one output layer. Each layer contains a set of $M_l$ neurons (where $l \in \{1, \ldots, L\}$) with Rectified Linear Unit (ReLU) activation functions. ReLU activation functions play an important role in the current advances in deep neural networks~\cite{krizhevsky2012imagenet}. For such neural network architecture, the neural network controller $u^{(t)} = f_{\text{NN}}(d(x^{(t)}))$ can be written as: 
\begin{align}
h^{1(t)} &= \max\left(0, \; W^0 d(x^{(t)}) + w^0 \right), \notag \\ 
h^{2(t)} &= \max \left(0, \; W^1 h^{1(t)} + w^1\right), \notag  \\ 
& \quad \vdots  \notag\\ 
h^{L(t)} &= max \left(0, \; W^{L-1} h^{L-1(t)} + w^{L-1}\right),	\notag  \\
u^{(t)} &= W^L h^{L(t)}	+ w^{L},
\label{eq:controller}    
\end{align}
where $W^l \in \R^{M_{i} \times M_{l-1}}$ and $w^l \in \R^{M_l}$ are the pre-trained weights and bias vectors of the neural network which are determined during the training phase.




\subsection{Robot Trajectories and Safety Specifications}
The trajectories of the robot whose dynamics are described by~\eqref{eq:dyn} when controlled by the neural network controller~\eqref{eq:image_func_nonlinear}-\eqref{eq:controller} starting from the initial condition $x_0 = x^{(0)}$ is denoted by $\eta_{x_0}: \N \to \R^n$ such that $\eta_{x_0}(0) = x_0$. A trajectory $\eta_{x_0}$ is said to be safe whenever the robot position does not collide with any of the obstacles at all times.
\begin{definition}[Safe Trajectory]
A robot trajectory $\eta_{x_0}$ is called safe if:
$$\zeta(\eta_{x_0}(t)) \not\in \mathcal{O}_i \qquad \qquad \forall \mathcal{O}_i \in \mathcal{O}, \; \forall t \in \N.$$	
\end{definition}

%
%

Using the previous definition, we now define the problem of verifying the system-level safety of the neural network controlled system as follows:
\begin{problem}
\label{prob:nnverification}
    Consider the autonomous robot whose dynamics are governed by~\eqref{eq:dyn} which is controlled by the neural network controller described by~\eqref{eq:controller} which processes LiDAR images described by~\eqref{eq:image_func_nonlinear}-\eqref{eq:image}. Compute the set of safe initial conditions $\mathcal{X}_{\text{safe}} \subseteq \mathcal{X}$ such that any trajectory $\eta_{x_0}$ starting from $x_0 \in \mathcal{X}_{\text{safe}}$ is safe.
    
\end{problem}

\begin{figure*}[!ht]
 \resizebox{.99\textwidth}{!}{
    \includegraphics[height=0.2\textwidth]{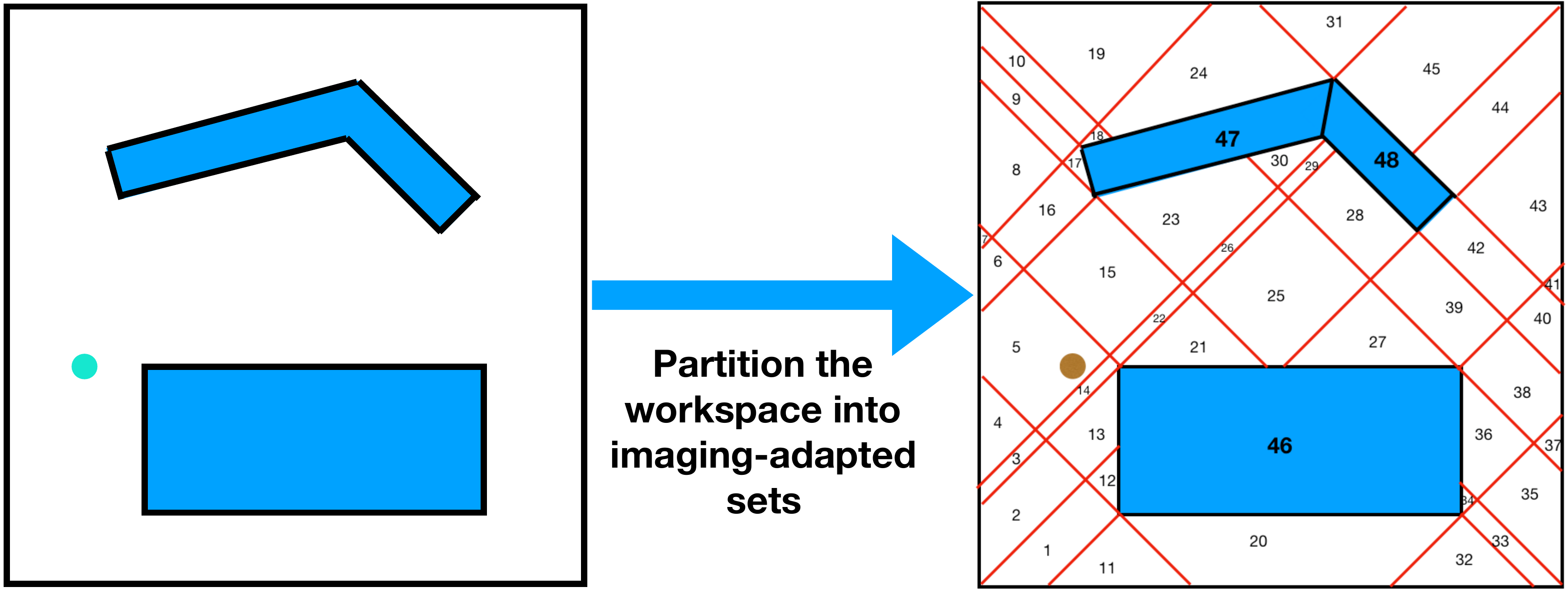}
    \includegraphics[height=0.2\textwidth]{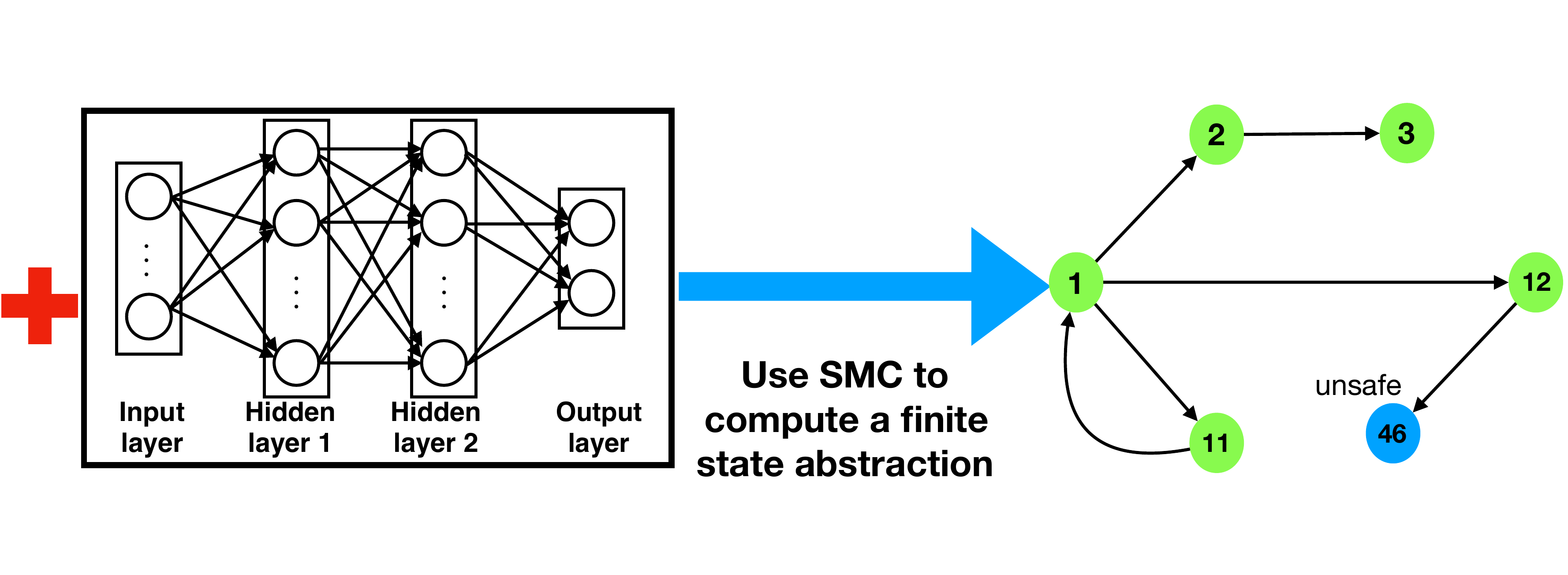}
    \includegraphics[height=0.19\textwidth]{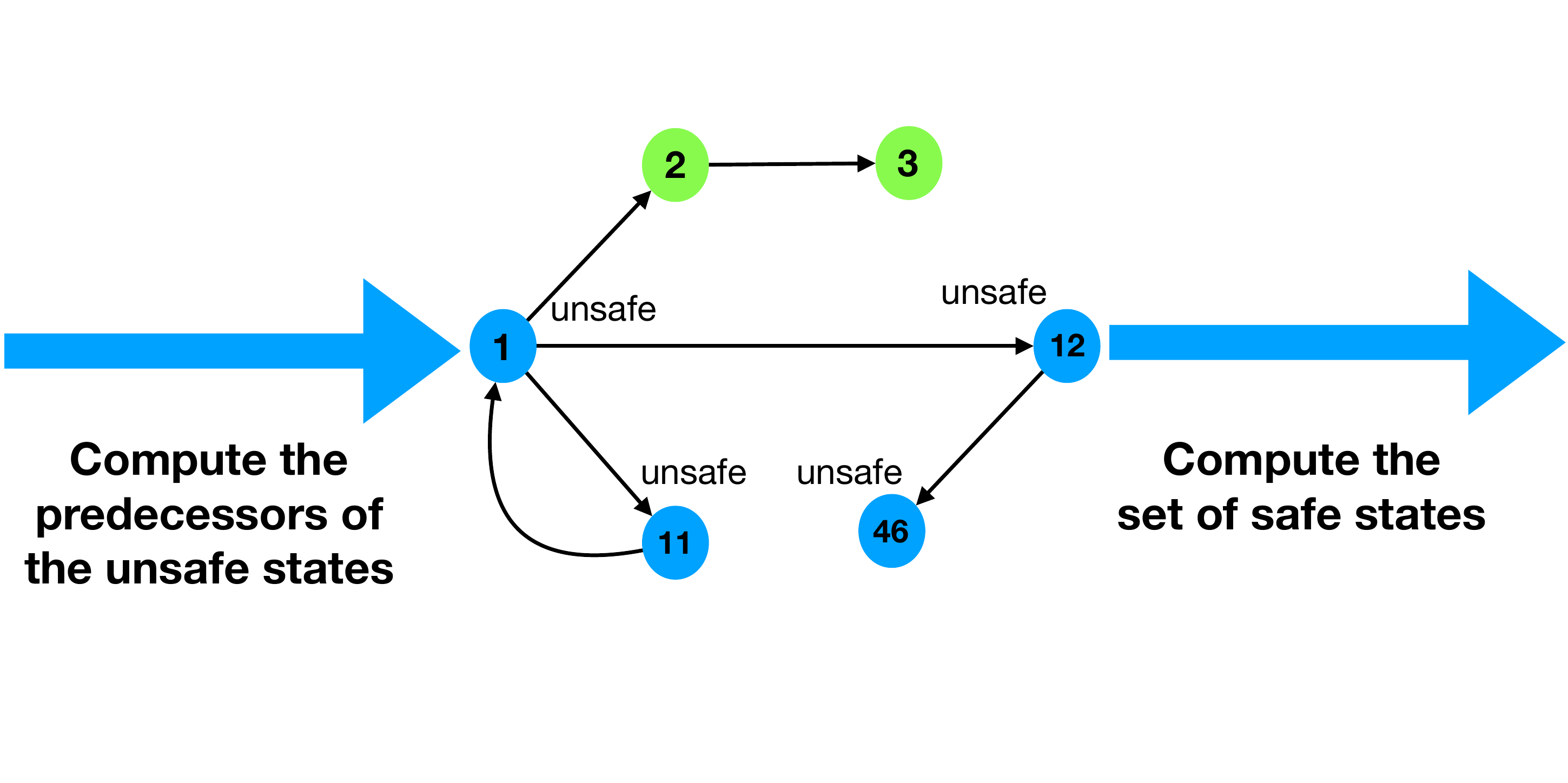}
    \includegraphics[height=0.2\textwidth]{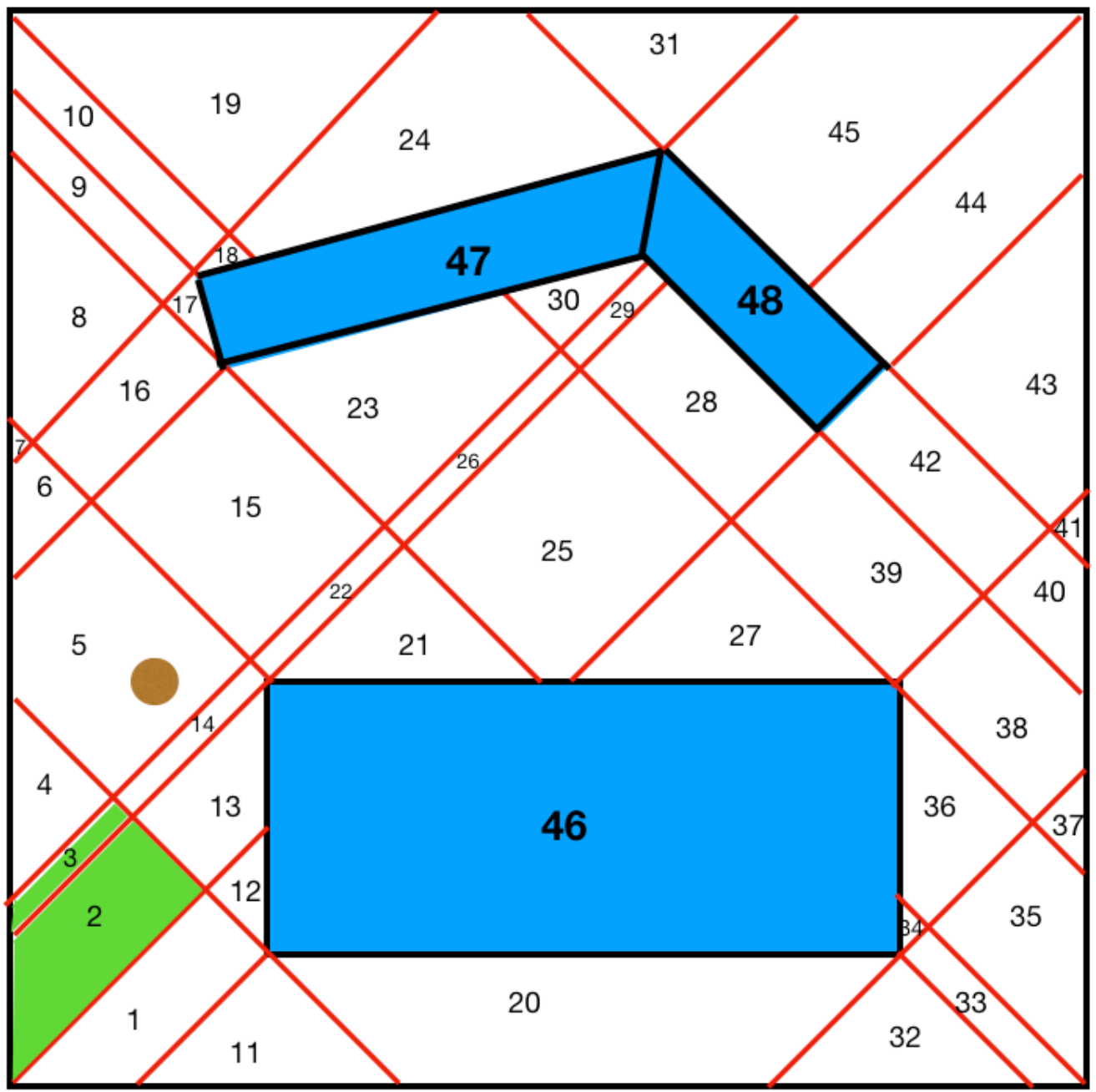} 
}    \vspace{-3mm}
    \caption{Pictorial representation of the proposed framework. \vspace{-3mm}}
    \label{fig:framework}
\end{figure*}

\section{Framework}

Before we describe the proposed framework, we need to briefly recall the following set of definitions capturing the notion of a system and relations between different systems.
\begin{definition}
	An autonomous system $\mathcal{S}$ is a pair $(X,\delta)$ consisting of a set of states $X$ and a set-valued map $\delta: X \rightrightarrows X$ representing the transition function. A system $\mathcal{S}$ is finite if $X$ is finite. A system $S$ is deterministic if $\delta$ is single-valued map and is non-deterministic if not deterministic.
\end{definition}

\begin{definition}
	Consider a deterministic system $\mathcal{S}_a = (X_a, \delta_a)$ and a non-deterministic $\mathcal{S}_b = (X_b, \delta_b)$. A relation $Q \subseteq X_a \times X_b$ is a simulation relation from $\mathcal{S}_a$ to $\mathcal{S}_b$, and we write $\mathcal{S}_a \preccurlyeq_{Q} \mathcal{S}_b$, if the following conditions are satisfied:
	\begin{enumerate}
		\item for every $x_a \in X_a$ there exists $x_b \in X_b$ with $(x_a,x_b) \in Q$,
		\item for every $(x_a,x_b) \in Q$ we have that $x_a' = \delta_a(x_a)$ in $S_a$ implies the existence of $x_b' \in \delta_b(x_b)$ in $S_b$ satisfying $(x_a', x_b') \in Q$.
	\end{enumerate} 
\end{definition}


Using the previous two definitions, we describe our approach as follows. As pictorially shown in Figure~\ref{fig:framework},
given the autonomous robot 
system $\mathcal{S}_{\text{NN}} = (\mathcal{X}, \delta_{\text{NN}})$, where $\delta_{\text{NN}}: x \mapsto A x + B f_{\text{NN}}(d(x))$, our objective is to compute a finite state abstraction (possibly non-deterministic) $\mathcal{S}_\mathcal{F} = ( \mathcal{F}, \delta_{\mathcal{F}} )$ of $\mathcal{S}_{\text{NN}}$ such that there exists a simulation relation from $\mathcal{S}_{\text{NN}}$ to $\mathcal{S}_\mathcal{F}$, i.e., $\mathcal{S}_{\text{NN}} \preccurlyeq_{Q} \mathcal{S}_\mathcal{F}$. This finite state abstraction $\mathcal{S}_\mathcal{F}$ will be then used to check the safety specification.




%
The first difficulty in computing the finite state abstraction $\mathcal{S}_\mathcal{F}$ is the nonlinearity in the relation between the robot position $\zeta(x)$ and the LiDAR observations as captured by equation~\eqref{eq:image_func_nonlinear}. However, we notice that we can partition the workspace based on the laser angles $\theta_1,\ldots,\theta_N$ along with the vertices of the polytopic obstacles such that the map $d$ (which maps the robot position to the processed observations) is an affine map. Therefore, as summarized in Algorithm~\ref{alg:framework}, the first step is to compute such partitioning $\mathcal{W}^*$ of the workspace (\textsc{WKSP-PARTITION}, line~\ref{line:wksppartition} in Algorithm~\ref{alg:framework}). While \textsc{WKSP-PARTITION} focuses on partitioning the workspace $\W$, one need to partition the remainder of the state space $\mathcal{X}$ (\textsc{STATE-SPACE-PARTITION}, line~\ref{line:statepartition} in Algorithm~\ref{alg:framework}) to compute the finite set of abstract states $\mathcal{F}$ along with the simulation relation $Q$ that maps between states in $\mathcal{X}$ and the corresponding abstract states in $\mathcal{F}$, and vice versa. 

Unfortunately, the number of partitions grows exponentially in the number of lasers $N$ and the number of vertices of the polytopic obstacles. To harness this exponential growth, we compute an aggregate-partitioning $\mathcal{W}^\prime$ using only few laser angles (called primary lasers and denoted by $\theta_p$). The resulting aggregate-partitioning $\mathcal{W}^\prime$ would contain a smaller number of partitions such that each partition in $\mathcal{W}^\prime$ represents multiple partitions in $\mathcal{W}$. Similarly, we can compute a corresponding aggregate set of states $\mathcal{F}^\prime$ as:
	\begin{align*}
		s' = \{ s \in \mathcal{F} \; \vert \; \exists x \in \mathcal{W}^\prime,  (x,s) \in Q \}	
	\end{align*}
	where each aggregate state $s'$ is a set representing multiple states in $\mathcal{F}$.
Whenever possible, we will carry out our analysis using the aggregated-partitioning  $\mathcal{W}^\prime$ (and $\mathcal{F}^\prime$) and use the fine-partitioning $\mathcal{W}$ only if deemed necessary. Details of the workspace partitioning and computing the corresponding affine maps representing the LiDAR imaging function are given in Section~\ref{sec:partition}.


\begin{algorithm}[t]
    \caption{\textsc{Verify-NN}($\mathcal{X}, \delta_{\text{NN}}$)}
    \label{alg:framework}
    {\small
    \begin{algorithmic}[1]
    	\STATE \textbf{Step 1: Partition the workspace}
        \STATE $(\W^*, \mathcal{W}^\prime)$ = \textsc{WKSP-PARTITION}($\W, \mathcal{O}, \theta_p, \theta_p$) \label{line:wksppartition}~\\~\\
        \STATE \textbf{Step 2: Compute the finite state abstraction $\mathcal{S}_\mathcal{F}$}
    	\STATE \textbf{Step 2.1: Compute the states of $\mathcal{S}_\mathcal{F}$ }        
        \STATE $(\mathcal{F}, \mathcal{F}^\prime, Q)$ = \textsc{STATE-SPACE-PARTITON}($\W^*, \mathcal{W}^\prime$) \label{line:statepartition} 
        \FOR{each $s$ and $s'$ in $\mathcal{F}$} \label{line:prepDelta1}
        	\STATE $\delta_{\mathcal{F}}.\textsc{ADD-TRANSITION}(s,s')$ \label{line:prepDelta2}
        \ENDFOR        	

        \STATE \textbf{Step 2.2: Pre-process the neural network}        
        \FOR{each $s$ and $s'$ in $\mathcal{F}$} \label{line:preprocess1}
        \STATE $\mathcal{X}_s$ = $\{x \in \mathcal{X} \; \vert \; (x,s) \in Q \}$        
        \STATE $CE_s$ = \textsc{PRE-PROCESS}($\mathcal{X}_s, \delta_{\text{NN}}$)   \label{line:preprocess2}      
        \ENDFOR        	

		\STATE \textbf{Step 2.3: Compute the transition map $\delta_{\mathcal{F}}$}        
        \FOR{each $s$ in $\mathcal{F}$ and $s'$ in $\mathcal{F}^\prime$ where $s \not \in s'$} \label{line:delta1}
        	\STATE $\mathcal{X}_s$ = $\{x \in \mathcal{X} \; \vert \; (x,s) \in Q \}$ 
        	\STATE $\mathcal{X}_{s'}$ = $\{x \in \mathcal{X} \; \vert \; (x,s^*) \in Q, \; \forall s^* \in s'\}$
        	\STATE \textsc{Status} = \textsc{CHECK-FEASIBILITY}($\mathcal{X}_s, \mathcal{X}_{s'}, \delta_{\text{NN}}, CE_s$)
        	\IF {\textsc{Status} == \texttt{INFEASIBLE}}
        		\FOR{each $s^\star$ in $s'$}
        		    \STATE $\delta_{\mathcal{F}}.\textsc{REMOVE-TRANSITION}(s,s^\star)$ \label{line:delta2}
        		\ENDFOR
        	\ELSE
        		\FOR{each $s^\star$ in $s'$} \label{line:delta3}
        			\STATE $\mathcal{X}_{s^\star}$ = $\{x \in \mathcal{X} \; \vert \; (x,s^*) \in Q \}$
        			\STATE \textsc{Status} = \texttt{CHECK-FEASIBILITY}($\mathcal{X}_s, \mathcal{X}_{s^\star}, \delta_{\text{NN}}, CE_s$)
        			\IF {\textsc{Status} == \texttt{INFEASIBLE}}
        				\STATE $\delta_{\mathcal{F}}.\textsc{REMOVE-TRANSITION}(s,s^\star)$ \label{line:delta4}
        			\ENDIF
        		\ENDFOR
        	\ENDIF
             
            	
        \ENDFOR~\\~\\
        \STATE \textbf{Step 3: Compute the safe set}
        \STATE \textbf{Step 3.1: Mark the abstract states corresponding to obstacles and workspace boundary as unsafe}
		\begin{align*}
 		\mathcal{F}^0_{\text{unsafe}} = \{s \in \mathcal{F} \; \vert \; &\exists x \in \mathcal{X}: (x,s) \in Q, \; \zeta(x) \in \mathcal{O}_i, \mathcal{O}_i \in  \mathcal{O} \}
	 	\end{align*} \label{line:safe1}
	 	\STATE \textbf{Step 3.2: Iteratively compute the predecessors of the abstract unsafe states}
		\STATE \textsc{Status} = \texttt{FIXED-POINT-NOT-REACHED}
		\WHILE{\textsc{Status} == \texttt{FIXED-POINT-NOT-REACHED}}
			\STATE $\mathcal{F}^k_{\text{unsafe}} = \mathcal{F}^{k-1}_{\text{unsafe}} \cup \textsc{Pre}(\mathcal{F}^{k-1}_{\text{unsafe}})$
			\IF{$\mathcal{F}^k_{\text{unsafe}} == \mathcal{F}^{k-1}_{\text{unsafe}}$}
				\STATE \textsc{Status} = \texttt{FIXED-POINT-REACHED}
			\ENDIF
		\ENDWHILE
		\STATE $\mathcal{F}_{\text{safe}} = \mathcal{F}\setminus\mathcal{F}_{\text{unsafe}}$
        \STATE \textbf{Step 3.3: Compute the set of safe states}		
		\STATE $\mathcal{X}_{\text{safe}} = \{x \in \mathcal{X} \; \vert \; \exists s \in \mathcal{F}_{\text{safe}} : (x,s) \in Q\}$ \label{line:safe2}
%
        \STATE \textbf{Return} $\mathcal{X}_\text{safe}$
    \end{algorithmic}     
    }
\end{algorithm}

The state transition map $\delta_{\mathcal{F}}$ is computed as follows. First, we assume a transition exists between any two states $s$ and $s'$ in $\mathcal{F}$ (line~\ref{line:prepDelta1}-~\ref{line:prepDelta2} in Algorithm~\ref{alg:framework}). Next, we start eliminating unnecessary transitions. We observe that regions in the workspace that are adjacent or within some vicinity are more likely to force the need of transitions between their corresponding abstract states. Similarly, regions in the workspace that are far from each other are more likely to prohibit transitions between their corresponding abstract states.   Therefore, in an attempt to reduce the number of computational steps in our algorithm, we check the transition feasibility between a state $s \in \mathcal{F}$ and an aggregate state $s' \in \mathcal{F}^\prime$. If our algorithm (\texttt{CHECK-FEASIBILITY}) asserted that the neural network $\delta_{\text{NN}}$ prohibits the robot from transitioning between the regions corresponding to $s$ and $s'$ (denoted by $\mathcal{X}_s$ and $\mathcal{X}_{s'}$, respectively), then we conclude that no transition in $\delta_{\mathcal{F}}$ is feasible between the abstract state $s$ and all the abstract states $s^\star$ in $s'$ (lines~\ref{line:delta1}-\ref{line:delta2} in Algorithm~\ref{alg:framework}). This leads to a reduction in the number of state pairs that need to be checked for transition feasibility. Conversely, if our algorithm (\texttt{CHECK-FEASIBILITY}) asserted that the neural network $\delta_{\text{NN}}$ allows for a transition between the regions corresponding to $s$ and $s'$, then we proceed by checking the transition feasibility between the state $s$ and all the states $s^*$ contained in the  aggregate state $s^*$ (lines~\ref{line:delta3}-\ref{line:delta4} in Algorithm~\ref{alg:framework}).


Checking the transition feasibility (\texttt{CHECK-FEASIBILITY}) between two abstract states entail reasoning about the robot dynamics, the neural network, along with the affine map representing the LiDAR imaging computed from the previous workspace partitioning. While the robot dynamics is assumed linear, the imaging function is affine; the technical difficulty lies in reasoning about the behavior of the neural network controller. Thanks to the ReLU activation functions in the neural network, we can encode the problem of checking the transition feasibility between two regions as formula $\varphi$, called monotone Satisfiability Modulo Convex (SMC) formula~\cite{shoukry2017smc,shoukry2018smc}, over Boolean and convex constraints representing, respectively, the ReLU phases and the dynamics, the neural network weights, and the imaging constraints. 
In addition to using SMC solver to check the transition feasibility (\texttt{CHECK-FEASIBILITY}) between abstract states, it will be used also to perform some pre-processing of the neural network function $\delta_{\text{NN}}$ (lines~\ref{line:preprocess1}-\ref{line:preprocess2} in Algorithm~\ref{alg:framework}) which is going to speed up the process of checking the the transition feasibility. Details of the SMC encoding are given in Section~\ref{sec:smc}.

Once the finite state abstraction $\mathcal{S}_\mathcal{F}$ and the simulation relation $Q$ is computed, the next step is to partition the finite states $\mathcal{F}$ into a set of unsafe states $\mathcal{F}_{\text{unsafe}}$ and a set of safe states $\mathcal{F}_{\text{unsafe}}$ using the following fixed-point computation:
\begin{align*}
	\mathcal{F}^k_{\text{unsafe}} &= 
	\begin{cases}
		\{s \in \mathcal{F} \; \vert \; \exists x \in \mathcal{X}: (x,s) \in Q, \\
		\qquad \qquad \zeta(x) \in \mathcal{O}_i, \mathcal{O}_i \in  \mathcal{O} \} & k = 0\\
		\mathcal{F}^{k-1}_{\text{unsafe}} \cup \textsc{Pre}(\mathcal{F}^{k-1}_{\text{unsafe}}) & k > 0
	\end{cases}\\
	\mathcal{F}_{\text{unsafe}} &= \lim_{k \to \infty} \mathcal{F}^k_{\text{unsafe}}\\
	\mathcal{F}_{\text{safe}} &= \mathcal{F}\setminus\mathcal{F}_{\text{unsafe}}.
\end{align*}
where the $\mathcal{F}^0_{\text{unsafe}}$ represents the abstract state corresponding to the obstacles and workspace boundaries while $\mathcal{F}^k_{\text{unsafe}}$ with $k > 0$ represents all the states that can reach $\mathcal{F}^0_{\text{unsafe}}$ in $k$-steps where:
$$\textsc{Pre}(s) = \{s' \in \mathcal{F} \; \vert \; s \in \delta_{\mathcal{F}}(s') \}. $$
The remaining abstract states are then marked as the set of safe states $\mathcal{F}_{\text{safe}}$. Finally, we can compute the set of safe states $\mathcal{X}_{\text{safe}}$ as:
$$ \mathcal{X}_{\text{safe}} = \{x \in \mathcal{X} \; \vert \; \exists s \in \mathcal{F}_{\text{safe}} : (x,s) \in Q\}.$$
These computations are summarized in lines~\ref{line:safe1}-\ref{line:safe2} in Algorithm~\ref{alg:framework}.





\section{Imaging-Adapted Workspace Partitioning} \label{sec:partition}

We start by introducing the notation of the important geometric objects. We denote by $\textsc{Ray}(w,\theta)$ the ray originated from a point $w \in \W$ in the direction $\theta$, i.e.:
$$\textsc{Ray}(w,\theta) = \{w' \in \W \; \vert \; \atantwo(w' - w) = \theta \}. $$
Similarly, we denote by $\textsc{Line}(w_1,w_2)$ the line segment between the points $w_1$ and $w_2$, i.e.:
$$\textsc{Line}(w_1,w_2) = \{w' \in \W \; \vert \; w' = \nu w_1 + (1 - \nu) w_2, \; 0 \le \nu \le 1 \}.$$
For a convex polytope $P \subseteq \W$, we denote by $\textsc{Vert}(P)$,  its set of vertices and by $\textsc{Edge}(P)$ its set of line segments representing the edges of the polyhedron. 
%

\subsection{Imaging-Adapted Partitions}
\begin{figure}[!t]
	\centering
    \includegraphics[width=0.8\columnwidth]{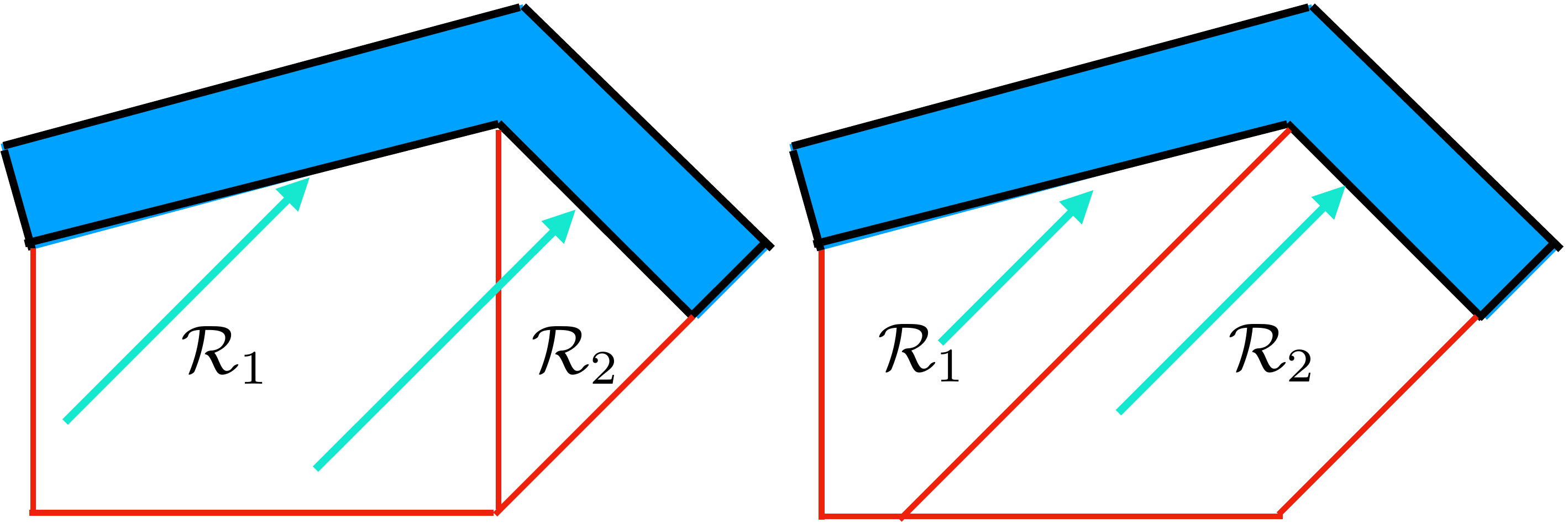}
    \caption{(left) A partitioning of the workspace that is \emph{not} imaging-adapted. Within region $\mathcal{R}_1$, the LiDAR ray (cyan arrow) intersects different obstacles edges depending on the robot position. (right) A partitioning of the workspace that is imaging-adapted. For both  regions $\mathcal{R}_1$ and $\mathcal{R}_2$, the LiDAR ray (cyan arrow) intersects the same obstacle edge regardless of the robot position.}
    \label{fig:sets}
\end{figure}

The basic idea behind our algorithm is to partition the workspace into a set of polytypic sets (or regions) such that for each region $\mathcal{R}$ the LiDAR rays will intersect the same obstacle/workspace edge regardless of the robot positions $\zeta(x) \in \mathcal{R}$. 
To formally characterize this property, let $\mathcal{O}^\star = \bigcup_{\mathcal{O}_i \in \mathcal{O}} \mathcal{O}_i$ be the set of all points $z$ in the workspace in which an obstacle or workspace boundary exists. 
Consider a workspace partition $\mathcal{R} \subseteq \W$ and a robot position $\zeta(x)$ that lies inside this partition, i.e., $\zeta(x) \in \mathcal{R}$. The intersection between the $k$th LiDAR laser beam $\textsc{Ray}(\zeta(x),\theta_k)$ and $\mathcal{O}^\star$ is a unique point characterized as:
	\begin{align}
		z_{k,\zeta(x)}(\mathcal{R}) &= \argmin_{z \in \W} \Vert z - \zeta(x) \Vert_2 \quad \text{s.t.} \quad z \in \textsc{Ray}(\zeta(x),\theta_k) \cap \mathcal{O}^\star.
		\label{eq:zkzeta}
	\end{align}
	By sweeping $\zeta(x)$ across the whole region $\mathcal{R}$, we can characterize the set of all possible intersection points as:
\begin{align}
		\mathcal{L}_k(\mathcal{R}) = \bigcup_{\zeta(x) \in \mathcal{R}} z_{k,\zeta(x)}(\mathcal{R}).
		\label{eq:L}
\end{align}
Using the set $\mathcal{L}_k(\mathcal{R})$ described above, we define the notion of imaging-adapted set as follows.
\begin{definition}
	A set $\mathcal{R} \subset \W$ is said to be an imaging-adapted partition if the following property holds:
	\begin{align}
		\mathcal{L}_k(\mathcal{R}) \text{ is a line segment} \quad \forall k \in \{1,\ldots,N\}.
		\label{eq:linesegment}
\end{align}
\end{definition}
Figure~\ref{fig:sets} shows concrete examples of imaging-adapted partitions. Imaging-adapted partitions enjoys the following property:
\begin{lemma}
	Consider an imaging-adapted partition $\mathcal{R}$ with corresponding sets $\mathcal{L}_1(\mathcal{R}), \ldots, \mathcal{L}_N(\mathcal{R})$. The LiDAR imaging function $d : \mathcal{R} \rightarrow \R^{2N}$ is an affine function of the form:
	\begin{align}
		d_k(\zeta(x)) &= P_{k,\mathcal{R}} \zeta(x) + Q_{k,\mathcal{R}}, \qquad d = (d_1, \ldots, d_N)
		\label{eq:dkfinal}
	\end{align}
	for some constant matrices $P_{k,\mathcal{R}}$ and vectors $Q_{k,\mathcal{R}}$ that depend on the region $\mathcal{R}$ and the LiDAR angle $\theta_k$.
	\label{lemma:regions}
\end{lemma}

\subsection{Partitioning the Workspace}
Motivated by Lemma~\ref{lemma:regions}, our objective is to design an algorithm that can partition the workspace $\W$ into a set of imaging-adapted partitios.  As summarized in Algorithm~\ref{alg:partition}, our algorithm starts by computing a set of line segments $\mathcal{G}$ that will be used to partition the workspace (lines~\ref{line:segments1}-\ref{line:segments2} in Algorithm~\ref{alg:partition}). This set of line segments $\mathcal{G}$ are computed as follows. First, we define the set $\mathcal{V}$ as the one that contains all the vertices of the workspace and the obstacles, i.e., $\mathcal{V} = \bigcup_{\mathcal{O}_i \in \mathcal{O}}\textsc{Vert}(\mathcal{O}_i)$. Next, we consider rays originating from all the vertices in $\mathcal{V}$ and pointing in the opposite direction of the angles $\theta_1,\ldots,\theta_N$. By intersecting these rays with the obstacles and picking the closest intersection points, we acquire the line segments $\mathcal{G}$ that will be used to partition the workspace. In other words, $\mathcal{G}$ is computed as: 
\begin{align}
\mathcal{G}_k &= \{\textsc{Line}(v, z) \; \vert \; v \in \mathcal{V},  
	\; z = \argmin_{z \in \textsc{Ray}(v, \theta_k + \pi) \cap \mathcal{O}^\star} \Vert z - v \Vert_2 \}\nonumber\\ 
\mathcal{G} &= \bigcup_{k = 1}^N 	\mathcal{G}_k \label{eq:G}
\end{align}
Thanks to the fact that the vertices $v$ are fixed, finding the intersection between $\textsc{Ray}(v, \theta_k + \pi)$ and $\mathcal{O}^\star$ is a standard ray-polytope intersection problem which can be solved efficiently~\cite{berg2008computational}.

The next step is to compute the intersection points $\mathcal{P}$ between the line segments $\mathcal{G}$ and the edges of the obstacles $\mathcal{E} = \bigcup_{\mathcal{O}_i \in \mathcal{O}} \textsc{Edge}(\mathcal{O}_i)$. A naive approach will be to consider all combinations of line segments in $\mathcal{G} \cup \mathcal{E}$ and test them for intersection. Such approach is combinatorial and would lead to an execution time that is exponential in the number of laser angles and vertices of obstacles. Thanks to the advancements in the literature of computational geometry, such intersection points can be computed efficiently using the \textsc{Plane-Sweep} algorithm~\cite{berg2008computational}. The plane-sweep algorithm simulates the process of sweeping a line downwards over the plane. The order of the line segments $\mathcal{G} \cup \mathcal{E}$ from left to right as they intersect the sweep line is stored in a data structure called the sweep-line status. Only segments that are adjacent in the horizontal ordering need to be tested for intersection. Though the sweeping process can be visualized as continuous, the plane-sweep algorithm sweeps only the values in which the endpoints of segments in $\mathcal{G} \cup \mathcal{E}$, which are given beforehand, and the intersection points, which are computed on the fly. To keep track of the endpoints of segments in $\mathcal{G} \cup \mathcal{E}$ and the intersection points, we use a balanced binary search tree as data structure to support insertion, deletion, and searching in $O(log \; n)$ time, where $n$ is number of elements in the data structure.

The final step is to use the line segments $\mathcal{G} \cup \mathcal{E}$ and their intersection points $\mathcal{P}$, discovered by the plane-sweep algorithm, to compute the workspace partitions. 
 	To that end, consider the undirected planar graph whose vertices are the intersection points $\mathcal{P}$ and whose edges are $\mathcal{G} \cup \mathcal{E}$, denoted by $\textsc{Graph}(\mathcal{P}, \mathcal{G} \cup \mathcal{E})$. The workspace partitions are equivalent to finding subgraphs of $\textsc{Graph}(\mathcal{P}, \mathcal{G} \cup \mathcal{E})$ such that each subgraph contains only one simple cycle \footnote{A cycle in an undirected graph is called simple when no repetitions of vertices and edges is allowed within the cycle.}. To find these simple cycles, we use a modified Depth-First-Search algorithm in which it starts from a vertex in the planar graph and then traverses the graph by considering the rightmost turns along the vertices of the graph. Finally, the workspace partition is computed as the convex hull of all the vertices in the computed simple cycles. It follows directly from the fact that each region is constructed from the vertices of a simple cycle that there exists no line segment in  $\mathcal{G} \cup \mathcal{E}$ that intersects with the interior of any region, i.e., for any workspace partition $\mathcal{R}$, the following holds:
 		\begin{align}
 			\text{int}(\mathcal{R}) \cap e = \emptyset \qquad \forall e \in \mathcal{G} \cup \mathcal{E}			
 			\label{eq:Rproperty}
 		\end{align}
 	This process is summarized in lines~\ref{line:partition1}-\ref{line:partition2} in Algorithm~\ref{alg:partition}. 
An important property of the regions determined by Algorithm~\ref{alg:partition} is stated by the following proposition.

\begin{proposition}
    \label{prop:lidar_config}
    Consider a partition $\mathcal{R}$ computed by Algorithm~\ref{alg:partition} and satisfies~\eqref{eq:Rproperty}. The following property holds for any LiDAR ray with angle $\theta_k$:
    $$ \exists e \in \mathcal{E}: \quad \mathcal{L}_k(\mathcal{R}) \subseteq e$$
    where $\mathcal{L}_k(\mathcal{R})$ defined in~\eqref{eq:L}.
    In other words, the LiDAR ray with angle $\theta_k$ intersects the same obstacle edge regardless of the robot position.
\end{proposition}

We conclude this section by stating our first main result, quantifying the correctness and complexity of Algorithm~\ref{alg:partition}.
\begin{theorem}
\label{Th:imaging}
    Given a workspace with polytopic obstacles and a set of laser angles $\theta_1,\ldots,\theta_N$, then Algorithm~\ref{alg:partition} computes the partitioning $\mathcal{R}_1, \ldots, \mathcal{R}_r$ such that:
    \begin{enumerate}
    	\item $\W = \bigcup_{i = 1}^r \mathcal{R}_i$,
    	\item $\mathcal{R}_i \text{ is an imaging-adapted partition } \quad \forall i = 1,\ldots, r$,
    	\item $d : \mathcal{R}_i \rightarrow \R^{2N} \text{ is affine } \qquad \qquad \qquad \forall i = 1,\ldots, r$.
    \end{enumerate}
    Moreover, the time complexity of Algorithm~\ref{alg:partition} is $O(M\log\;M + I\log\;M)$, where $M = |\mathcal{G} \cup \mathcal{E}|$ is cardinality of $\mathcal{G} \cup \mathcal{E}$, and $I$ is number of intersection points between segments in $\mathcal{E} \cup \mathcal{E}$.
    
\end{theorem}    


\begin{algorithm}[t]
    \caption{\textsc{WKSP-PARTITION} ($\W, \mathcal{O}, \theta, \theta_p$ )}
    \label{alg:partition}
    {\small
    \begin{algorithmic}[1]
        \STATE \textbf{Step 1: Generate partition segments} \label{line:segments1}
        \STATE $
        \mathcal{O}^\star = \bigcup_{\mathcal{O}_i \in \mathcal{O}} \mathcal{O}_i, \quad
        \mathcal{V} = \bigcup_{\mathcal{O}_i \in \mathcal{O}}\textsc{Vert}(\mathcal{O}_i), \quad 
        \mathcal{E} = \bigcup_{\mathcal{O}_i \in \mathcal{O}} \textsc{Edge}(\mathcal{O}_i)
        $
		\FOR{$k \in \{1,\ldots,N\}$}
			\STATE Use a ray-polygon intersection algorithm to compute:
			$$\qquad \quad \mathcal{G}_k = \{\textsc{Line}(v, z) \; \vert \; v \in \mathcal{V},  
	\; z = \argmin_{z \in \textsc{Ray}(v, \theta_k + \pi) \cap \mathcal{O}^\star} \Vert z - v \Vert_2 \}$$ 
		\ENDFOR
		\STATE $\mathcal{G} = \bigcup_{k \in \theta} 	\mathcal{G}_k, \qquad \mathcal{G}' = \bigcup_{k \in \theta_p} \mathcal{G}_k $ \label{line:segments2}~\\~\\

		\STATE \textbf{Step 2: Compute intersection points}
		\STATE $\mathcal{P} = \textsc{PLANE-SWEEP}(\mathcal{G} \cup \mathcal{E}), \qquad\mathcal{P}' = \textsc{PLANE-SWEEP}(\mathcal{G}' \cup \mathcal{E})$~\\~\\
		
		\STATE \textbf{Step 3: Construct the  partitions} \label{line:partition1}
		\STATE \textsc{Cycles} = \textsc{Find-Vertices-Of-Simple-Cycle}($\textsc{Graph}(\mathcal{P}, \mathcal{G} \cup \mathcal{E})$) 
		\STATE \textsc{Cycles}' = \textsc{Find-Vertices-Of-Simple-Cycle} ($\textsc{Graph}(\mathcal{P}', \mathcal{G}' \cup \mathcal{E})$).
        \FOR{$c \in \textsc{Cycles}$}
        	\STATE $\mathcal{R} = \textsc{Convex-Hull}(c)$
        	\STATE $\mathcal{W}^\star.\textsc{ADD}(\mathcal{R})$
        \ENDFOR
        \FOR{$c \in \textsc{Cycles}'$}
        	\STATE $\mathcal{R}' = \textsc{Convex-Hull}(c)$
        	\STATE $\mathcal{W}'.\textsc{ADD}(\mathcal{R}')$ \label{line:partition2}
        \ENDFOR
        \STATE \textbf{Return} $\mathcal{W}^\star, \mathcal{W}'$

    \end{algorithmic}  
    }
\end{algorithm}

\section{Computing the Finite State Abstraction} \label{sec:smc}


Once the workspace is partitioned into imaging-adapted regions $\mathcal{W}^\star = \{\mathcal{R}_1,\ldots,\mathcal{R}_r\}$ and the corresponding imaging function is identified, the next step is to compute the finite state transition abstraction $\mathcal{S}_\mathcal{F} = (\mathcal{F}, \delta_{\mathcal{F}})$ of the closed loop system along with the simulation relation $Q$. The first step is to define the state space $\mathcal{F}$ and its relation to $\mathcal{X}$.
To that end, we start by computing a partitioning of the state space $\mathcal{X}$ that respects $\mathcal{W}^\star$. For the sake of simplicity, we consider $\mathcal{X} \subset \R^{n}$ that is $n$-orthotope, i.e., there exists constants $\underline{x}_i , \overline{x}_i \in \R, i = 1,\ldots, n$ such that:
	$$ \mathcal{X} = \{x \in \R^n \; \vert \;  \underline{x}_i \le  x_i < \overline{x}_i, \quad i = 1,\ldots, n\} $$
Now, given a discretization parameter $\epsilon \in \R^{+}$, we define the state space $\mathcal{F}$ as:
\begin{align}
	\mathcal{F} = \{(k_1, k_3, \ldots, k_n) &\in \N^{n-1} \; \vert \; 
			1 \le k_1 \le r, \nonumber \\
			&1 \le k_i \le \frac{\overline{x}_i - \underline{x}_i}{\epsilon}, i = 3, \ldots, n \}
	\label{eq:F_definition}		
\end{align}
where $r$ is the number of regions in the partitioning $\mathcal{W}^\star$. In other words, the parameter $\epsilon$ is used to partition the state space into $\epsilon$ hyper-cubes. A state $s \in \mathcal{F}$ represents the index of a region in $\mathcal{W}^\star$ followed by the indices identifying a hypercube in the remaining $n-2$ dimensions. Note that for the simplicity of notation, we assume that $\overline{x}_i - \underline{x}_i$ is divisible by $\epsilon$ for all $i = 1,\ldots,n$. We now define the relation $Q \subseteq \mathcal{X} \times \mathcal{F}$ as:
\begin{align}
	Q = \{(x,s) \in \mathcal{X} \times \mathcal{F} \; \vert \; &s = (k_1, k_3, \ldots, k_n),  x = (\zeta(x), x_3, \ldots, x_n), \nonumber\\
	& \zeta(x) \in \mathcal{R}_{k_1}, \underline{x}_i + \epsilon (k_i - 1) \le x_i < \underline{x}_i + \epsilon k_i,\nonumber\\
	& i = 3, \ldots, n \label{eq:Q_definition}
	 \}.
\end{align}

Finally, we define the state transition function $\delta_{\mathcal{F}}$ of $\mathcal{S}_\mathcal{F}$ as follows:
\begin{align}
	&(k_1', k_3', \ldots k_n') \in \delta_{\mathcal{F}}((k_1, k_3, \ldots k_n)) \text{ if } \nonumber\\
	&\quad \exists x = (\zeta(x), x_3,\ldots,x_n) \in \mathcal{R}_{k_1}, \underline{x}_i + \epsilon (k_i - 1) \le x_i < \underline{x}_i + \epsilon k_i, \nonumber\\
	&\quad \; x' = (\zeta(x'), x_3',\ldots,x_n') \in \mathcal{R}_{k_1'}, \underline{x}_i + \epsilon (k_i' - 1) \le x_i' < \underline{x}_i + \epsilon k_i',\nonumber \\
	&\quad \text{s.t.} \quad x' = Ax + Bf_{NN}(d(x)).
	\label{eq:deltaF_definition}
 \end{align}
It follows from the definition of $\delta_{\mathcal{F}}$ in~\eqref{eq:deltaF_definition} that checking the transition feasibility between two states $s$ and $s'$ is equivalent to searching for a robot initial and goal states along with a LiDAR image that will force the neural network controller to generate an input that moves the robot between the two states while respecting the robots dynamics. In the reminder of this section, we focus on solving this feasibility problem.

\subsection{SMC Encoding of NN}
We translate the problem of checking the transition feasibility in $\delta_{\mathcal{F}}$ into a feasibility problem over a monotone SMC formula~\cite{shoukry2017smc,shoukry2018smc} as follows. We introduce the Boolean indicator variables $b_j^l$ with $l = 1,\ldots,L$ and $j = 1,\ldots, M_l$ (recall that $L$ represents the number of layers in the neural network, while $M_l$ represents the number of neurons in the $l$th layer). These Boolean variables represents the phase of each ReLU, i.e., an asserted $b_j^l$ indicates that the output of the $j$th ReLU in the $l$th layer is $h_j^l = (W^{l-1} h^{l-1} + w^{l-1})_j$ while a negated $b_j^l$ indicates that $h_j^l = 0$. Using these Boolean indicator variables, we encode the problem of checking the transition feasibility between two states $s = (k_1,k_3,\ldots,k_n)$ and $s' = (k_1',k_3',\ldots,k_n')$ as:
\begin{align}
			&\exists \; x, x' \in \R^n, u \in \R^m, d \in \R^{2N}, \\
					   &\qquad(b^l, h^l, t^l) \in \B^{M_l}\times \R^{M_l} \times \R^{M_l}, \quad l \in \{1,\ldots,L\} \nonumber\\
			& \text{subject to:} \nonumber\\
			& \zeta(x) \in \mathcal{R}_{k_1} \; \land  \; \underline{x}_i + \epsilon (k_i - 1) \le x_i < \overline{x}_i + \epsilon k_i, \; i = 3,\ldots,n \label{eq:encodingFrom}\\
			\land & \zeta(x') \in \mathcal{R}_{k_1'} \land  \underline{x}_i + \epsilon (k_i' - 1) \le x_i' < \overline{x}_i + \epsilon k_i', \; i = 3,\ldots,n \label{eq:encodingTo}\\
			\land & \; x' = A x + B u \label{eq:encodingDyn}\\
			\land &  \; d_k = P_{k,\mathcal{R}_{k_1}} \zeta(x) + Q_{k,\mathcal{R}_{k_1}}, \qquad k = 1, \ldots, N \label{eq:encodingImaging}\\
			\land & \; \bigg( t^{1} = W^0 d + w^0 \bigg) \; \land  \; \bigg( \bigwedge_{l = 2}^L t^{l} = W^{l-1} h^{l-1} + w^{l} \bigg) \label{eq:encodingNN1}\\
			\land  &   \; \bigg( u = W^{L} h^{L} + w^{L} \bigg) \label{eq:encodingNN2}\\
			\land &  \; \bigwedge_{l = 1}^{L} \bigwedge_{j = 1}^{M_j}  b^l_j \rightarrow \left[ \left( h^l_j = t_j^{l} \right) \land \left(t_j^{l} \ge 0 \right) \right] 
			\label{eq:encodingNN3}\\
			\land &   \; \bigwedge_{l = 1}^{L} \bigwedge_{j = 1}^{M_j}  \lnot b^l_j \rightarrow \left[ \left(h^l_j = 0 \right) \land \left( t_j^{l} < 0 \right)	\right] \label{eq:encodingNN4}
\end{align}
where~\eqref{eq:encodingFrom}-\eqref{eq:encodingTo} encodes the state space partitions corresponding to the states $s$ and $s'$; \eqref{eq:encodingDyn} encodes the dynamics of the robot; \eqref{eq:encodingImaging} encodes the imaging function that maps the robot position into LiDAR image; \eqref{eq:encodingNN1}-\eqref{eq:encodingNN4} encodes the neural network controller that maps the LiDAR image into a control input.

Compared to Mixed-Integer Linear Programs (MILP), monotone SMC formulas avoid using encoding heuristics like big-M encodings which leads to numerical instabilities. SMC decision procedures follow an iterative approach combining efficient Boolean Satisfiability (SAT) solving with numerical convex programming. When applied to the encoding above, at each iteration the SAT solver generates a candidate assignment for the ReLU indicator variables $b_j^l$. The correctness of these assignments are then checked by solving the corresponding set of convex constraints. If the convex program turned to be infeasible, indicating a wrong choice of the ReLU indicator variables, the SMC solver will identify the set of ``Irreducible Infeasible Set'' (IIS) in the convex program to provide the most succinct explanation of the conflict. This IIS will be then fed back to the SAT solver to prune its search space and provide the next assignment for the ReLU indicator variables. SMC solvers was shown to better handle problems (compared with MILP solvers) for problems with relatively large number of Boolean variables~\cite{shoukry2018smc}.

\subsection{Pruning Search Space By Pre-processing}
While a neural network with $M$ ReLUs would give rise to $2^M$ combinations of possible assignments to the corresponding Boolean indicator variables, we observe that several of those combinations are infeasible   for each workspace region. In other words, the LiDAR imaging function along with the workspace region enforces some constraints on the inputs to the neural network which in turn enforces constraints on the subsequent layers. By performing pre-processing on each of the workspace regions, we can discover those constraints and augment it to the SMC encoding~\eqref{eq:encodingFrom}-\eqref{eq:encodingNN4} to prune several combinations of assignments of the ReLU indicator variables.

To find such constraints, we consider an SMC problem with the fewer constraints~\eqref{eq:encodingFrom},~\eqref{eq:encodingImaging}-\eqref{eq:encodingNN4}. By iteratively solving the reduced SMC problem and recording all the IIS conflicts produced by the SMC solver, we can compute a set of counter-examples that are unique for each region. By iteratively invoking the SMC solver while adding previous counter-examples as constraints until the problem is no longer satisfiable, we compute the set $\mathcal{R}\textsc{-Conflicts}$ which represents all the counter-examples for region $\mathcal{R}$. Finally, we add the following constraint:
\begin{align}
	\bigwedge_{c \in \mathcal{R}\textsc{-Conflicts}} c
	\label{eq:encodingCE}
\end{align}
to the original SMC encoding~\eqref{eq:encodingFrom}-\eqref{eq:encodingNN4} to prune the set of possible assignments to the ReLU indicator variables. In Section~\ref{sec:results}, we show that pre-processing would result in an order of magnitude reduction in the execution time.

%
%
%
%
%
%
%
%
%
%
%

\subsection{Correctness of Algorithm~\ref{alg:framework}}
We end our discussion with the following results which asserts the correctness of the whole framework described in this paper. We first start by establishing the correctness of computing the finite abstraction $\mathcal{S}_{\mathcal{F}}$ along with the simulation relation $Q$ as follows:
\begin{proposition}
\label{prop:simulation}
	Consider the finite state system $\mathcal{S}_\mathcal{F} = (\mathcal{F}, \delta_\mathcal{F})$ where $\mathcal{F}$ is defined by~\eqref{eq:F_definition} and $\delta_\mathcal{F}$ is defined by~\eqref{eq:deltaF_definition} and computed by means of solving the SMC formula~\eqref{eq:encodingFrom}-\eqref{eq:encodingCE}. Consider also the system $\mathcal{S}_{\text{NN}} = (\mathcal{X}, \delta_{\text{NN}})$ where $\delta_{\text{NN}} = x \mapsto Ax + B f_{\text{NN}}(d(x))$. For the relation $Q$ defined in~\eqref{eq:Q_definition}, the following holds:
	$\qquad \mathcal{S}_{\text{NN}} \preccurlyeq_Q \mathcal{S}_\mathcal{F}$.
\end{proposition}

Recall that Algorithm~\ref{alg:framework} applies standard reachability analysis on $\mathcal{S}_\mathcal{F}$ to compute the set of unsafe states. It follows directly from the correctness of the simulation relation $Q$ established above that our algorithm computes an over-approximation of the set of unsafe states, and accordingly an under-approximation of the set of safe states. This fact is captured by the following result that summarizes the correctness of the proposed framework:
\begin{theorem}
	\label{Th:correcteness}
	Consider the safe set $\mathcal{X}_{\text{safe}}$ computed by Algorithm~\ref{alg:framework}. Then any trajectory $\eta_x$ with $\eta_x(0) \in \mathcal{X}_{\text{safe}}$ is a safe trajectory.
\end{theorem}
While Theorem~\ref{Th:correcteness} establishes the correctness of the proposed framework in Algorithm~\ref{alg:framework}, two points needs to be investigated namely (i) complexity of Algorithm~\ref{alg:framework} and (ii) maximality of the set $\mathcal{X}_{\text{safe}}$. Although Algorithm~\ref{alg:partition} computes the imaging-adapted partitions efficiently (as shown in Theorem~\ref{Th:imaging}), analyzing a neural network with ReLU activation functions is shown to be NP-hard. Exacerbating the problem, Algorithm~\ref{alg:framework} entails analyzing the neural network a number of times that is exponential in the number of partition regions. In Section~\ref{sec:results}, we experiment the efficiency of using SMC decision procedures to harness this computational complexity. As for the maximality of the computed $\mathcal{X}_{\text{safe}}$ set, we note that Algorithm~\ref{alg:framework} is not guaranteed to search for the maximal $\mathcal{X}_{\text{safe}}$.



\section{Results}  \label{sec:results}

We implemented the proposed verification framework as described by Algorithm~\ref{alg:framework} on top of the SMC solver named \textit{SATEX}~\cite{satex}.
All experiments were executed on an Intel Core i7 2.5-GHz processor with 16 GB of memory. 

\subsection{Scalability of the Workspace Partitioning Algorithm:}
As the first step of our verification framework, imaging-adapted workspace partitioning is tested for numerical stability with increasing number of laser angles and obstacles. 
Table~\ref{tab:partition} summarizes the scalability results in terms of the number of computed regions and the execution time grows as the number of LiDAR lasers and obstacle vertices increase. Thanks to adopting well-studied computational geometry algorithms, our partitioning process takes less than $1.5$ minutes for the scenario where a LiDAR scanner is equipped with 298 lasers (real-world LiDAR scanners are capable of providing readings from 270 laser angles).


\begin{table}
    \caption{Scalability results for the \textsc{WKSP-PARTITION} Algorithm\vspace{-3mm}}
    \begin{center}
    \resizebox{.9\columnwidth}{!}{
    \begin{tabular}{|c|c|c|c|c|}
    \hline
    \textbf{Number of} & \textbf{Number of} & \textbf{Number of} &\textbf{Time} \\ 
    \textbf{Vertices}  & \textbf{Lasers}    & \textbf{regions}   &\textbf{[s]}  \\
    \hline\hline
         & 8 	& 111   & 0.0152    \\ \cline{2-4}
    8 	 & 38 	& 1851  & 0.3479    \\ \cline{2-4}
         & 118 	& 17237 & 5.5300 	 \\ \hline
         & 8 	& 136   & 0.0245    \\ \cline{2-4}
    10 	 & 38 	& 2254  & 0.4710    \\ \cline{2-4}
         & 118  & 20343 & 6.9380    \\ \hline
         & 8 	& 137   & 0.0275    \\ \cline{2-4}
     	 & 38 	& 2418  & 0.5362    \\ \cline{2-4}
    12   & 120 	& 23347 & 8.0836    \\ \cline{2-4}
         & 218 	& 76337 & 37.0572   \\ \cline{2-4}
         & 298 	& 142487    & 86.6341   \\ \hline
\end{tabular}
}
\end{center}
\label{tab:partition}
\vspace{-5mm}
\end{table}

\subsection{Computational Reduction Due to Pre-processing}
The second step is to pre-process the neural network. In particular, we would like to answer the following question: given a partitioned workspace, how many ReLU assignments are feasible in each region, and if any, what is the execution time to find them out. Recall that a ReLU assignment is feasible if there exist a robot position and the corresponding LiDAR image that will lead to that particular ReLU assignment. 

Thanks to the IIS counterexample strategy, we can find all feasible ReLU assignments in pre-processing. 
Our first observation is that the number of feasible assignments is indeed much smaller compared to the set of all possible assignments. As shown in Table~\ref{tab:preprocess-scalability}, for a neural network with a total of 32 neurons, only 11 ReLU assignments are feasible (within the region under consideration). Comparing this number to $2^{32} = 4.3E9$ possibilities of ReLU assignments, we conclude that pre-processing is very effective in reducing the search space by several orders of magnitude.


Furthermore, we conducted an experiment to study the scalability of the proposed pre-processing for an increasing number of ReLUs. To that end, we fixed one choice of workspace regions while changing the neural network architecture. The execution time, the number of generated counterexamples, along with the number of feasible ReLU assignments are given in Table~\ref{tab:preprocess-scalability}.
For the case of neural networks with one hidden layer, our implementation of the counterexample strategy is able to find feasible ReLU assignments for a couple of hundreds of neurons in less than 4 minutes. In general, the number of counterexamples, and hence feasible ReLU assignments, and execution time grows with the number of neurons.
However, the number of neurons is not the only deciding factor. Our experiments show that the depth of the network plays a significant role in affecting the scalability of the proposed algorithms. For example, comparing the neural network with one hidden layer and a hundred neurons per layer versus the network with two layers and fifty neurons per layer we notice that both networks share the same number of neurons. Nevertheless, the deeper network resulted in one order of magnitude increase regarding the number of generated counterexamples and one order of magnitude increase in the corresponding execution time. Interestingly, both of the architectures share a similar number of feasible ReLU assignments.
 In other words, similar features of the neural network can be captured by fewer counterexamples whenever the neural network has fewer layers. This observation can be accounted for the fact that counterexamples that correspond to ReLUs in early layers are more powerful than those involves ReLUs in the later layers of the network. 

In the second part of this experiment, we study the dependence of the number of feasible ReLU assignments on the choice of the workspace region. To that end, we fix the architecture of the neural network to one with $2$ hidden layers and $40$ neurons per layer. Table~\ref{tab:preprocess-regions} reports the execution time, the number of counterexamples, and the number of feasible ReLU assignments across different regions of the workspace. In general, we observe that the number of feasible ReLU assignments increases with the size of the region.
%
%
%
%

\begin{table}
    \caption{Execution time of the SMC-based pre-processing as a function of the neural network architecture. \vspace{-3mm}}
    \begin{center}
    \resizebox{.99\columnwidth}{!}{
    \begin{tabular}{|c|c|c|c|c|c|}
    \hline
    \textbf{Number}     & \textbf{Total} & \textbf{Number of}   &\textbf{Number of} &\textbf{Time} \\ 
    \textbf{of Hidden} & \textbf{Number}   & \textbf{feasibile} &\textbf{Counter}   &\textbf{[s]}  \\
    \textbf{Layers}              & \textbf{of Neurons}             & \textbf{ReLU Assignments} &\textbf{Examples} &\textbf{} \\
    \hline\hline
        & 32    & 11    & 60    & 2.7819    \\ \cline{2-5}
        & 72    & 31    & 183	& 11.4227   \\ \cline{2-5}
        & 92    & 58    & 265	& 18.4807   \\ \cline{2-5}
        & 102   & 68    & 364	& 43.2459   \\ \cline{2-5}
        & 152   & 101   & 540	& 78.3015   \\ \cline{2-5}
        & 172   & 146   & 778	& 104.4720  \\ \cline{2-5}
        & 202   & 191   & 897	& 227.2357  \\ \cline{2-5}
    1   & 302   & 383   & 1761	& 656.3668  \\ \cline{2-5}
        & 402   & 730   & 2614	& 1276.4405 \\ \cline{2-5}
       & 452   & 816   & 4325	& 1856.0418 \\ \cline{2-5}
        & 502   & 1013  & 3766	& 2052.0574 \\ \cline{2-5}
        & 552   & 1165  & 4273	& 4567.1767 \\ \cline{2-5}
        & 602   & 1273  & 5742  & 6314.4890 \\ \cline{2-5}
        & 652   & 1402  & 5707	& 7166.3059 \\ \cline{2-5}
        & 702   & 1722  & 6521	& 8813.1829 \\ \hline
        & 22 	& 3     & 94    & 1.3180    \\ \cline{2-5}
        & 42 	& 19    & 481   & 10.9823   \\ \cline{2-5}
        & 62 	& 35    & 1692  & 53.2246   \\ \cline{2-5}
        & 82 	& 33    & 2685  & 108.2584  \\ \cline{2-5}
    2   & 102 	& 58    & 5629  & 292.7412  \\ \cline{2-5}
        & 122 	& 71    & 9995  & 739.4883  \\ \cline{2-5}
        & 142 	& 72    & 18209 & 2098.0220 \\ \cline{2-5}
        & 162 	& 98    & 34431 & 6622.1830 \\ \cline{2-5}
        & 182 	& 152   & 44773 & 12532.8552\\ \hline
        & 32    & 5     & 319   & 5.7227    \\ \cline{2-5}
    3   & 47    & 7     & 5506  & 148.8727  \\ \cline{2-5}
        & 62    & 45    & 72051 & 12619.5353\\ \hline
    4   & 22    & 9     & 205   & 10.4667    \\ \cline{2-5}
        & 42    & 5     & 1328  & 90.1148   \\ \hline
\end{tabular}       
}
\end{center}
\label{tab:preprocess-scalability}
\vspace{-5mm}
\end{table}

\begin{table}
    \caption{Execution time of the SMC-based pre-processing as a function of the workspace region. \vspace{-3mm}}
    \begin{center}
    \resizebox{.99\columnwidth}{!}{
    \begin{tabular}{|c|c|c|c|c|}
    \hline
    \textbf{Region}        & \textbf{Number of}   &\textbf{Number of} &\textbf{Time} \\ 
    \textbf{Index} & \textbf{feasibile} &\textbf{Counter}   &\textbf{[s]}  \\
    \textbf{}              & \textbf{ReLU Assignments} &\textbf{Examples} &\textbf{} \\
    \hline\hline
    A2-R3   & 33    & 2685  & 108.2584  \\ \cline{1-4}
    A14-R1  & 55    & 4925  & 215.8251  \\ \cline{1-4}
    A13-R3  & 7     & 1686  & 69.4158   \\ \cline{1-4}
    A1-R1   & 25    & 2355  & 99.2122   \\ \cline{1-4}
    A7-R1   & 26    & 3495  & 139.3486  \\ \cline{1-4}
    A12-R2  & 3     & 1348  & 54.4548   \\ \cline{1-4}
    A15-R3  & 25    & 3095  & 121.7869  \\ \cline{1-4}
    A19-R1  & 38    & 4340  & 186.6428  \\ \hline
\end{tabular}       
}
\end{center}
\label{tab:preprocess-regions}
\vspace{-5mm}
\end{table}

\subsection{Transition Feasibility}
Following our verification streamline, the next step is to compute the transition function of the finite-state abstraction $\delta_\mathcal{F}$, i.e., check transition feasibility between regions.
 Table~\ref{tab:transition-scalability} shows performance comparison between our proposed strategy that uses counterexamples obtained from pre-processing and SMC encoding without preprocessing.
 We observe that SMC encodings empowered by counterexamples, generated through the pre-processing phase, scales more favorably compared to the ones that do not take counterexamples into account leading to 2-3 orders of magnitude reduction in the execution time. Moreover, and thanks to the pre-processing counter-examples, we observe that checking transition feasibility becomes less sensitive to changes in the neural network architecture as shown in  Table~\ref{tab:transition-scalability}.
 
 



\begin{table}
    \caption{Performance of the SMC-based encoding for computing $\delta_\mathcal{F}$ as a function of the neural network (timeout = 1 hour/) \vspace{-3mm}}
    \begin{center}
    \resizebox{.99\columnwidth}{!}{
    \begin{tabular}{|c|c|c|c|c|}
    \hline
    \textbf{Number of}     & \textbf{Total Number} & \textbf{Time [s]}          &\textbf{Time [s]}         \\ 
    \textbf{Hidden Layers} & \textbf{of Neurons}   & \textbf{(Exploit Counter} &\textbf{(Without Counter} \\
    \textbf{}              & \textbf{}             & \textbf{Examples)}         &\textbf{Examples)}        \\
    \hline\hline

        & 82    & 0.5056    & 50.1263   \\ \cline{2-4}
        & 102   & 7.1525    & timeout   \\ \cline{2-4}
    1   & 112   & 12.524    & timeout   \\ \cline{2-4}
        & 122   & 18.0689   & timeout   \\ \cline{2-4}
        & 132   & 20.4095   & timeout   \\ \hline
        & 22    & 0.1056    & 15.8841   \\ \cline{2-4}
        & 42    & 4.8518    & timeout   \\ \cline{2-4}
        & 62    & 3.1510    & timeout   \\ \cline{2-4}
        & 82    & 2.6112    & timeout   \\ \cline{2-4}
    2   & 102   & 11.0984   & timeout   \\ \cline{2-4}
        & 122   & 3.8860    & timeout   \\ \cline{2-4}
        & 142   & 0.7608    & timeout   \\ \cline{2-4}
        & 162   & 2.7917    & timeout   \\ \cline{2-4}
        & 182   & 193.6693  & timeout   \\ \hline
        & 32    & 0.3884    & 388.549  \\ \cline{2-4}
    3   & 47    & 0.9034    & timeout  \\ \cline{2-4}
        & 62    & 59.393    & timeout  \\ \hline
\end{tabular}       
}
\end{center}
\label{tab:transition-scalability}
\vspace{-5mm}
\end{table}

%

\section{Conclusions}
We presented a framework to verify the safety of autonomous robots equipped with LiDAR scanners and controlled by neural network controllers. Thanks to the notion of imaging-adapted sets, we can partition the workspace to render the problem amenable to formal verification. Using SMC-encodings, we presented a framework to compute finite-state abstraction of the system that can be used to compute an under-approximation of the set of safe robot states. We demonstrated a pre-processing technique that generates a set of counterexamples which resulted in 2-3 orders of magnitude execution time reduction. Future work includes investigating further strategies for efficient generation of pre-processing counterexamples along with extending the proposed technique to account to uncertainties in the LiDAR scanner.

\bibliographystyle{IEEEtran}
\bibliography{biblio}

\begin{thebibliography}{10}
\providecommand{\url}[1]{#1}
\csname url@samestyle\endcsname
\providecommand{\newblock}{\relax}
\providecommand{\bibinfo}[2]{#2}
\providecommand{\BIBentrySTDinterwordspacing}{\spaceskip=0pt\relax}
\providecommand{\BIBentryALTinterwordstretchfactor}{4}
\providecommand{\BIBentryALTinterwordspacing}{\spaceskip=\fontdimen2\font plus
\BIBentryALTinterwordstretchfactor\fontdimen3\font minus
  \fontdimen4\font\relax}
\providecommand{\BIBforeignlanguage}[2]{{%
\expandafter\ifx\csname l@#1\endcsname\relax
\typeout{** WARNING: IEEEtran.bst: No hyphenation pattern has been}%
\typeout{** loaded for the language `#1'. Using the pattern for}%
\typeout{** the default language instead.}%
\else
\language=\csname l@#1\endcsname
\fi
#2}}
\providecommand{\BIBdecl}{\relax}
\BIBdecl

\bibitem{AccidentWiki}
Wikipedia, ``List of autonomous car fatalities,''
  \url{https://en.wikipedia.org/wiki/List_of_autonomous_car_fatalities}.

\bibitem{ferdowsi2018robust}
A.~Ferdowsi, U.~Challita, W.~Saad, and N.~B. Mandayam, ``Robust deep
  reinforcement learning for security and safety in autonomous vehicle
  systems,'' \emph{arXiv preprint}, 2018.

\bibitem{everitt2018agi}
T.~Everitt, G.~Lea, and M.~Hutter, ``{AGI} safety literature review,''
  \emph{arXiv preprint}, 2018.

\bibitem{charikar2017learning}
M.~Charikar, J.~Steinhardt, and G.~Valiant, ``Learning from untrusted data,''
  in \emph{Proceedings of the 49th Annual ACM SIGACT Symposium on Theory of
  Computing}.\hskip 1em plus 0.5em minus 0.4em\relax ACM, 2017, pp. 47--60.

\bibitem{steinhardt2017certified}
J.~Steinhardt, P.~W.~W. Koh, and P.~S. Liang, ``Certified defenses for data
  poisoning attacks,'' in \emph{Advances in Neural Information Processing
  Systems}, 2017, pp. 3520--3532.

\bibitem{munoz2017towards}
L.~Mu{\~n}oz-Gonz{\'a}lez, B.~Biggio, A.~Demontis, A.~Paudice, V.~Wongrassamee,
  E.~C. Lupu, and F.~Roli, ``Towards poisoning of deep learning algorithms with
  back-gradient optimization,'' in \emph{Proceedings of the 10th ACM Workshop
  on Artificial Intelligence and Security}.\hskip 1em plus 0.5em minus
  0.4em\relax ACM, 2017, pp. 27--38.

\bibitem{paudice2018label}
A.~Paudice, L.~Mu{\~n}oz-Gonz{\'a}lez, and E.~C. Lupu, ``Label sanitization
  against label flipping poisoning attacks,'' \emph{arXiv preprint}, 2018.

\bibitem{ruan2018global}
W.~Ruan, M.~Wu, Y.~Sun, X.~Huang, D.~Kroening, and M.~Kwiatkowska, ``Global
  robustness evaluation of deep neural networks with provable guarantees for l0
  norm,'' \emph{arXiv preprint}, 2018.

\bibitem{pei2017deepxplore}
K.~Pei, Y.~Cao, J.~Yang, and S.~Jana, ``Deepxplore: Automated whitebox testing
  of deep learning systems,'' in \emph{Proceedings of the 26th Symposium on
  Operating Systems Principles}.\hskip 1em plus 0.5em minus 0.4em\relax ACM,
  2017, pp. 1--18.

\bibitem{tian2017deeptest}
Y.~Tian, K.~Pei, S.~Jana, and B.~Ray, ``Deeptest: Automated testing of
  deep-neural-network-driven autonomous cars,'' \emph{arXiv preprint
  arXiv:1708.08559}, 2017.

\bibitem{wicker2018feature}
M.~Wicker, X.~Huang, and M.~Kwiatkowska, ``Feature-guided black-box safety
  testing of deep neural networks,'' in \emph{International Conference on Tools
  and Algorithms for the Construction and Analysis of Systems}.\hskip 1em plus
  0.5em minus 0.4em\relax Springer, 2018, pp. 408--426.

\bibitem{YouchengTesting2018}
Y.~Sun, X.~Huang, and D.~Kroening, ``Testing deep neural networks,''
  \emph{arXiv preprint}, 2018.

\bibitem{LeiDeepGauge2018}
L.~Ma, F.~Juefei{-}Xu, J.~Sun, C.~Chen, T.~Su, F.~Zhang, M.~Xue, B.~Li, L.~Li,
  Y.~Liu, J.~Zhao, and Y.~Wang, ``Deepgauge: Comprehensive and
  multi-granularity testing criteria for gauging the robustness of deep
  learning systems,'' \emph{arXiv preprint}, 2018.

\bibitem{Wang2018Testing}
J.~Wang, J.~Sun, P.~Zhang, and X.~Wang, ``Detecting adversarial samples for
  deep neural networks through mutation testing,'' \emph{arXiv preprint}, 2018.

\bibitem{LeiDeepMutation2018}
L.~Ma, F.~Zhang, J.~Sun, M.~Xue, B.~Li, F.~Juefei{-}Xu, C.~Xie, L.~Li, Y.~Liu,
  J.~Zhao, and Y.~Wang, ``Deepmutation: Mutation testing of deep learning
  systems,'' \emph{arXiv preprint}, 2018.

\bibitem{srisakaokul2018multiple}
S.~Srisakaokul, Z.~Wu, A.~Astorga, O.~Alebiosu, and T.~Xie,
  ``Multiple-implementation testing of supervised learning software,'' in
  \emph{Proceedings of the AAAI-18 Workshop on Engineering Dependable and
  Secure Machine Learning Systems (EDSMLS)}, 2018.

\bibitem{MengshiDeepRoad2018}
M.~Zhang, Y.~Zhang, L.~Zhang, C.~Liu, and S.~Khurshid, ``Deeproad: Gan-based
  metamorphic autonomous driving system testing,'' \emph{arXiv preprint}, 2018.

\bibitem{YouchengConcolic2018}
Y.~Sun, M.~Wu, W.~Ruan, X.~Huang, M.~Kwiatkowska, and D.~Kroening, ``Concolic
  testing for deep neural networks,'' \emph{arXiv preprint}, 2018.

\bibitem{dreossi2017compositional}
T.~Dreossi, A.~Donz{\'e}, and S.~A. Seshia, ``Compositional falsification of
  cyber-physical systems with machine learning components,'' in \emph{NASA
  Formal Methods Symposium}.\hskip 1em plus 0.5em minus 0.4em\relax Springer,
  2017, pp. 357--372.

\bibitem{tuncali2018simulation}
C.~E. Tuncali, G.~Fainekos, H.~Ito, and J.~Kapinski, ``Simulation-based
  adversarial test generation for autonomous vehicles with machine learning
  components,'' \emph{arXiv preprint arXiv:1804.06760}, 2018.

\bibitem{zhang2018two}
Z.~Zhang, G.~Ernst, S.~Sedwards, P.~Arcaini, and I.~Hasuo, ``Two-layered
  falsification of hybrid systems guided by monte carlo tree search,''
  \emph{IEEE Transactions on Computer-Aided Design of Integrated Circuits and
  Systems}, 2018.

\bibitem{kurd2003establishing}
Z.~Kurd and T.~Kelly, ``Establishing safety criteria for artificial neural
  networks,'' in \emph{International Conference on Knowledge-Based and
  Intelligent Information and Engineering Systems}.\hskip 1em plus 0.5em minus
  0.4em\relax Springer, 2003, pp. 163--169.

\bibitem{seshia2016towards}
S.~A. Seshia, D.~Sadigh, and S.~S. Sastry, ``Towards verified artificial
  intelligence,'' \emph{arXiv preprint}, 2016.

\bibitem{seshia2018formal}
S.~A. Seshia, A.~Desai, T.~Dreossi, D.~Fremont, S.~Ghosh, E.~Kim,
  S.~Shivakumar, M.~Vazquez-Chanlatte, and X.~Yue, ``Formal specification for
  deep neural networks,'' \emph{arXiv preprint}, 2018.

\bibitem{leikeAIsafety2017}
J.~Leike, M.~Martic, V.~Krakovna, P.~A. Ortega, T.~Everitt, A.~Lefrancq,
  L.~Orseau, and S.~Legg, ``{AI} safety gridworlds,'' \emph{arXiv preprint},
  2017.

\bibitem{leofante2018automated}
F.~Leofante, N.~Narodytska, L.~Pulina, and A.~Tacchella, ``Automated
  verification of neural networks: Advances, challenges and perspectives,''
  \emph{arXiv preprint}, 2018.

\bibitem{scheibler2015towards}
K.~Scheibler, L.~Winterer, R.~Wimmer, and B.~Becker, ``Towards verification of
  artificial neural networks,'' in \emph{Workshop on Methoden und
  Beschreibungssprachen zur Modellierung und Verifikation von Schaltungen und
  Systemen (MBMV)}, 2015, pp. 30--40.

\bibitem{katz2017reluplex}
G.~Katz, C.~Barrett, D.~L. Dill, K.~Julian, and M.~J. Kochenderfer, ``Reluplex:
  An efficient smt solver for verifying deep neural networks,'' in
  \emph{International Conference on Computer Aided Verification}.\hskip 1em
  plus 0.5em minus 0.4em\relax Springer, 2017, pp. 97--117.

\bibitem{ehlers2017formal}
R.~Ehlers, ``Formal verification of piece-wise linear feed-forward neural
  networks,'' in \emph{International Symposium on Automated Technology for
  Verification and Analysis}.\hskip 1em plus 0.5em minus 0.4em\relax Springer,
  2017, pp. 269--286.

\bibitem{bunel2018unified}
R.~Bunel, I.~Turkaslan, P.~H.~S. Torr, P.~Kohli, and M.~P. Kumar, ``A unified
  view of piecewise linear neural network verification,'' \emph{arXiv
  preprint}, 2018.

\bibitem{ruan2018reachability}
W.~Ruan, X.~Huang, and M.~Kwiatkowska, ``Reachability analysis of deep neural
  networks with provable guarantees,'' \emph{arXiv preprint}, 2018.

\bibitem{dutta2018output}
S.~Dutta, S.~Jha, S.~Sankaranarayanan, and A.~Tiwari, ``Output range analysis
  for deep feedforward neural networks,'' in \emph{NASA Formal Methods
  Symposium}.\hskip 1em plus 0.5em minus 0.4em\relax Springer, 2018, pp.
  121--138.

\bibitem{pulina2010abstraction}
L.~Pulina and A.~Tacchella, ``An abstraction-refinement approach to
  verification of artificial neural networks,'' in \emph{International
  Conference on Computer Aided Verification}.\hskip 1em plus 0.5em minus
  0.4em\relax Springer, 2010, pp. 243--257.

\bibitem{tjeng2017verifying}
V.~Tjeng and R.~Tedrake, ``Verifying neural networks with mixed integer
  programming,'' \emph{arXiv preprint arXiv:1711.07356}, 2017.

\bibitem{gehr2018ai}
T.~Gehr, M.~Mirman, D.~Drachsler-Cohen, P.~Tsankov, S.~Chaudhuri, and
  M.~Vechev, ``Ai 2: Safety and robustness certification of neural networks
  with abstract interpretation,'' in \emph{Security and Privacy (SP), 2018 IEEE
  Symposium on}, 2018.

\bibitem{xiang2017reachable}
W.~Xiang, H.-D. Tran, and T.~T. Johnson, ``Reachable set computation and safety
  verification for neural networks with relu activations,'' \emph{arXiv
  preprint arXiv:1712.08163}, 2017.

\bibitem{xiang2017survey}
W.~Xiang, P.~Musau, A.~A. Wild, D.~M. Lopez, N.~Hamilton, X.~Yang,
  J.~Rosenfeld, and T.~T. Johnson, ``Verification for machine learning,
  autonomy, and neural networks survey,'' \emph{arXiv preprint
  arXiv:1810.01989}, 2018.

\bibitem{kahn2017plato}
G.~Kahn, T.~Zhang, S.~Levine, and P.~Abbeel, ``Plato: Policy learning using
  adaptive trajectory optimization,'' in \emph{Robotics and Automation (ICRA),
  2017 IEEE International Conference on}.\hskip 1em plus 0.5em minus
  0.4em\relax IEEE, 2017, pp. 3342--3349.

\bibitem{krizhevsky2012imagenet}
A.~Krizhevsky, I.~Sutskever, and G.~E. Hinton, ``Imagenet classification with
  deep convolutional neural networks,'' in \emph{Advances in neural information
  processing systems}, 2012, pp. 1097--1105.

\bibitem{shoukry2017smc}
Y.~Shoukry, P.~Nuzzo, A.~L. Sangiovanni-Vincentelli, S.~A. Seshia, G.~J.
  Pappas, and P.~Tabuada, ``{SMC: Satisfiability Modulo Convex} optimization,''
  in \emph{Proceedings of the 20th International Conference on Hybrid Systems:
  Computation and Control (HSCC)}.\hskip 1em plus 0.5em minus 0.4em\relax ACM,
  2017, pp. 19--28.

\bibitem{shoukry2018smc}
------, ``Smc: Satisfiability modulo convex programming [40pt],''
  \emph{Proceedings of the IEEE}, vol. 106, no.~9, pp. 1655--1679, 2018.

\bibitem{berg2008computational}
M.~d. Berg, O.~Cheong, M.~v. Kreveld, and M.~Overmars, \emph{Computational
  geometry: algorithms and applications}.\hskip 1em plus 0.5em minus
  0.4em\relax Springer-Verlag TELOS, 2008.

\bibitem{satex}
\BIBentryALTinterwordspacing
(2018, Jun.) {SatEX Solver}. [Online]. Available:
  \url{https://yshoukry.bitbucket.io/SatEX/}
\BIBentrySTDinterwordspacing

\end{thebibliography}

\appendix
\section{Proof of Lemma~\ref{lemma:regions}} 

\begin{proof}
	Consider an arbitrary LiDAR laser with an angle $\theta_k$ and arbitrary robot position $\zeta(x) \in \mathcal{R}$. The LiDAR image $d_k$ can be written as:\vspace{-2mm}
	\begin{align}
		 d_k = z_{k,\zeta(x)}(\mathcal{R}) - \zeta(x)
		 \label{eq:dkproof}
	\end{align}
	where $z_{k,\zeta(x)}(\mathcal{R})$ is defined in~\eqref{eq:zkzeta}. 
	It follows from the fact that $\mathcal{R}$ is an imaging-adapted partition that the set $\mathcal{L}_k(\mathcal{R})$ is a line segment. Let $a_k,b_k \in \R^2$ be the vertices of this line segment, i.e., $(a_k,b_k) = \textsc{Vert}(\mathcal{L}_k(\mathcal{R}))$ and recall that $z_{k,\zeta(x)}(\mathcal{R})$ satisfies $z_{k,\zeta(x)}(\mathcal{R}) \in \mathcal{L}_k(\mathcal{R})$ and hence $z_{k,\zeta(x)}(\mathcal{R})$ lies on the line segment $\textsc{Line}(a_k,b_k)$. Therefore there exists a $\nu_k$ such that:
	\begin{align}
		z_{k,\zeta(x)}(\mathcal{R}) = (1- \nu_k) a_k + \nu b_k
		\label{eq:zproof}		
	\end{align}
	where $0 \le \nu_k \le 1$. It follows from the definition of $z_{k,\zeta(x)}(\mathcal{R})$ in~\eqref{eq:zkzeta} that $z_{k,\zeta(x)}(\mathcal{R})$ also lies on $\textsc{Ray}(\zeta(x),\theta_k)$ and hence:
	\begin{align}
		\tan(\theta_k) = \frac{z_2 - x_2}{z_1 - x_1}	,
		\label{eq:zproof2}
	\end{align}
	where $(z_1, z_2)$ are the two elements of $z_{k,\zeta(x)}(\mathcal{R}) \in \mathcal{R} \subset \R^2$ while $(x_1,x_2)$ are the corresponding two elements of $\zeta(x) \in \mathcal{R} \subset \R^2$. Substituting~\eqref{eq:zproof} in~\eqref{eq:zproof2} yields:
	\begin{align}
		\tan(\theta_k) = \frac{(1-\nu_k)a_2 + \nu_k b_2 - x_2}{(1-\nu_k)a_1 + \nu_k a_2 - x_1}	
		\label{eq:zproof3}
	\end{align}
	where $(a_1,a_2) = a_k$ and $(b_1,b_2) = b_k$ are the two elements of $a_k$ and $b_k$, respectively. By solving~\eqref{eq:zproof3} for $\nu_k$, we conclude that:
	\begin{align}
	\nu_k &= A_{\nu_k} \zeta(x) + b_{\nu_k},	 \label{eq:nu}\\
	A_{\nu_k} &= 
	\begin{cases}
 		\matrix{\frac{1}{b_2 - a_2} & 0} \qquad \qquad\qquad\qquad\qquad  \theta_k = \pi/2 \text{ or } 3\pi/2\\~\\
 		\matrix{\frac{\tan(\theta_k)}{a_2 - b_2 + (b_1 - a_1) \tan(\theta_k) - a_2} & \frac{1}{a_2 - b_2 + (b_1 - a_1) \tan(\theta_k)} } \\
 		\qquad\qquad\qquad\qquad\qquad\qquad\qquad\quad \text{otherwise},
 	\end{cases}
	\nonumber \\
	b_{\nu_k}&=  
	\begin{cases}
		0 & 	\qquad\qquad\qquad \theta_k = \pi/2 \text{ or } 3\pi/2 \\
		\frac{a_2 - a_2 \tan(\theta_k)}{a_2 - b_2 + (b_1 - a_1) \tan(\theta_k)} & \qquad\qquad\qquad\text{otherwise},
	\end{cases}	\nonumber 
	\end{align}
	where $A_{\nu_k}$ and $b_{\nu_k}$ are constants that depends on the values of the constants $a_k,b_k,$ and $\theta_k$.
	From~\eqref{eq:dkproof},\eqref{eq:zproof}, and~\eqref{eq:nu}, we conclude that:
	\begin{align}
		d_k(\zeta(x)) &= P_{k,\mathcal{R}} \zeta(x) + Q_{k,\mathcal{R}}
		\label{eq:dkfinal2}
	\end{align}
	with $P_{k,\mathcal{R}} = (b_k - a_k) (A - I)$ (where $I$ is the $2\times2$ identity matrix) and $Q_{k,\mathcal{R}} = a_k + b_{\nu_k}(b_k - a_k)$ are constants that depends on $a_k,b_k$ and $\theta_k$ form which we conclude that $d_k(\zeta(x))$ is affine. Note that we added the subscript $\mathcal{R}$ to $P_{k,\mathcal{R}}$ and $Q_{k,\mathcal{R}}$ to emphasize the face that these constant matrices depends on the region $\mathcal{R}$.  Since we picked $k$ arbitrary, we finally conclude that $d(\zeta(x))$ is also an affine function.
\end{proof}

\section{Proof of Proposition~\ref{prop:lidar_config}}
\begin{proof}
	We assume, for the sake of contradiction, that there exists two obstacle edge $\textsc{Line}(v_1,v_2),\textsc{Line}(w_1,w_2) \in \mathcal{E}$ with $(v_1,v_2) \ne (w_1,w_2)$ along with rays originating from points $p_1,p_2 \in  \mathcal{R}$ such that the intersection points:
	\begin{align*}
		z_1 &= \argmin_{z_1 \in \textsc{Ray}(p_1, \theta_k) \cap \mathcal{O}^\star} \Vert z_1 - p_1\Vert\\
		z_2 &= \argmin_{z_2 \in \textsc{Ray}(p_2, \theta_k) \cap \mathcal{O}^\star} \Vert z_2 - p_2\Vert	.	
	\end{align*}
	satisfy $z_1 \in \textsc{Line}(v_1,v_2)$ and $z_2 \in \textsc{Line}(w_1,w_2)$. 
	
	Now consider the set $P_1$ defined as follows:
	$$P_1 = \{ p \in \mathcal{R} \; \vert \; p \in \textsc{Ray}(z, \theta_k + \pi), \quad \forall z \in \textsc{Line}(v_1,v_2) \} $$
	It follows from the definition of $z_1$ that $p_1 \in P_1$. It also follows from the definition of $P_1$ that $P_1 \subseteq \mathcal{R}$. 
	Moreover, it follows from the definition of the set $\mathcal{E}$ along with the fact that $\textsc{Line}(v_1,v_2) \in \mathcal{E}$ that $v_1$ and $v_2$ satisfy $v_1,v_2 \in \mathcal{V}$. 
	It follows from the definition of the set $\mathcal{G}$ in~\eqref{eq:G} that it contains line segments from the rays originated at elements of the set $\mathcal{V}$. Hence, 	there exists $v_1', v_1'', v_2', v_2''$ such that the line segments $\textsc{Line}(v_1', v_1'')$ and $\textsc{Line}(v_2', v_2'')$ satisfy:
	\begin{align}
		&\textsc{Line}(v_1', v_1'') \subset \textsc{Ray}(v_1, \theta_k + \pi) \subset P_1 \subseteq \mathcal{R}, \qquad \textsc{Line}(v_1', v_1'') \in \mathcal{G} \label{eq:linev1}\\
		&\textsc{Line}(v_2', v_2'') \subset \textsc{Ray}(v_2, \theta_k + \pi) \subset P_1 \subseteq \mathcal{R}, \qquad \textsc{Line}(v_2', v_2'') \in \mathcal{G}		
		\label{eq:linev2}
	\end{align}
	However, it follows from~\eqref{eq:Rproperty} that line segments that are elements of $\mathcal{G}$ do not intersect the interior of $\mathcal{R}$. Hence: 
%
\begin{align}
&\left.\begin{aligned}
\textsc{Line}(v_1', v_1'') \subset \mathcal{R} \\
\textsc{Line}(v_1', v_1'') \cap \text{int}(\mathcal{R}) = \emptyset
\end{aligned}\right\rbrace
\Rightarrow
\textsc{Line}(v_1', v_1'') \subset \partial \mathcal{R} \label{eq:lines1contradiction}
\\
&\left.\begin{aligned}
\textsc{Line}(v_2', v_2'') \subset \mathcal{R} \\
\textsc{Line}(v_2', v_2'') \cap \text{int}(\mathcal{R}) = \emptyset
\end{aligned}\right\rbrace
\Rightarrow
\textsc{Line}(v_2', v_2'') \subset \partial \mathcal{R} \label{eq:lines2contradiction}
\end{align}

%
%
%
%

	Similarly, by considering $w_1,w_2, z_2$, we conclude that there exists line segments $\textsc{Line}(w_1', w_1'') \subset \textsc{Ray}(w_1, \theta_k + \pi)$ and $\textsc{Line}(w_2', w_2'') \subset \textsc{Ray}(v_2, \theta_k + \pi)$ are elements of $\mathcal{G}$ and satisfy: 
		\begin{align}
			&\textsc{Line}(w_1', w_1'') \subset \partial \mathcal{R}, \quad
			\textsc{Line}(w_2', w_2'') \subset \partial \mathcal{R}
			\label{eq:lines3contradiction}
		\end{align}
	
	It follows from Euclidean geometry that any polygon in $\R^2$ can have at maximum two edges that are ``parallel''. It also follows from~\eqref{eq:lines1contradiction}-\eqref{eq:lines3contradiction} that $\textsc{Line}(v_1', v_1''),\textsc{Line}(v_2', v_2''),\textsc{Line}(w_1', w_1''),$ and $\textsc{Line}(w_2', w_2'')$ are edges of $\mathcal{R}$. However, it follows from the definitions of the four line segments that they are subsets of rays that share the same angle, and hence they are all parallel. Hence we conclude that $\textsc{Line}(v_1',v_1'') = \textsc{Line}(w_1',w_1'')$ and $\textsc{Line}(v_2',v_2'') = \textsc{Line}(w_2',w_2'')$ from which it is direct to conclude that $(v_1,v_2) = (w_1,w_2)$, a contradiction.
\end{proof}

\section{Proof of Theorem~\ref{Th:imaging}}

\begin{proof}
	Property (1) follows from Proposition~\ref{prop:lidar_config} where (2) follows from Lemma~\ref{lemma:regions}. The complexity of the partitioning follows from the plane-sweep algorithm whose complexity is established in Theorem 2.4 in~\cite{berg2008computational}.
\end{proof}

\section{Proof of Proposition~\ref{prop:simulation}}
\begin{proof}
	It follows from Theorem~\ref{Th:imaging} that the LiDAR imaging is affine and the partitions $\mathcal{R}$ are convex and hence the encoding in~\eqref{eq:encodingFrom}-\eqref{eq:encodingCE} is indeed monotone SMC. The result then holds as a consequence of the correctness of the SMC decision procedure used to solve~\eqref{eq:encodingFrom}-\eqref{eq:encodingCE} which in turns entails the correctness of computing $\delta_{\mathcal{F}}$.
\end{proof}

%

\end{document}